\documentclass{article}

\usepackage{arxiv}

\usepackage[utf8]{inputenc} % allow utf-8 input
\usepackage[T1]{fontenc}    % use 8-bit T1 fonts
\usepackage{hyperref}       % hyperlinks
\usepackage{url}            % simple URL typesetting
\usepackage{booktabs}       % professional-quality tables
\usepackage{amsfonts}       % blackboard math symbols
\usepackage{nicefrac}       % compact symbols for 1/2, etc.
\usepackage{microtype}      % microtypography

\usepackage{graphicx}
\usepackage{sidecap}

\title{Alleviating catastrophic forgetting using context-dependent gating and synaptic stabilization}

\date{} 					% Or removing it

\author{
    Nicolas Y.~Masse\thanks{To whom correspondence may be addressed} \\
    Department of Neurobiology \\
    The University of Chicago \\
    Chicago, IL 60637 \\
    \texttt{masse@uchicago.edu} \\
    \And
    Gregory D.~Grant \\
    Department of Neurobiology \\
    The University of Chicago \\
    Chicago, IL 60637 \\
    \texttt{gdgrant@uchicago.edu} \\
    \AND
    David J.~Freedman\footnotemark[1] \\
    Department of Neurobiology \\
    The Grossman Institute for Neuroscience, \\ Quantitative Biology and Human Behavior \\
    The University of Chicago \\
    Chicago, IL 60637 \\
    \texttt{dfreedman@uchicago.edu} \\
}

\begin{document}
\maketitle

\begin{abstract}
Humans and most animals can learn new tasks without forgetting old ones. However, training artificial neural networks (ANNs) on new tasks typically causes them to forget previously learned tasks. This phenomenon is the result of "catastrophic forgetting", in which training an ANN disrupts connection weights that were important for solving previous tasks, degrading task performance. Several recent studies have proposed methods to stabilize connection weights of ANNs that are deemed most important for solving a task, which helps alleviate catastrophic forgetting. Here, drawing inspiration from algorithms that are believed to be implemented \textit{in vivo}, we propose a complementary method: adding a context-dependent gating signal, such that only sparse, mostly non-overlapping patterns of units are active for any one task. This method is easy to implement, requires little computational overhead, and allows ANNs to maintain high performance across large numbers of sequentially presented tasks, particularly when combined with weight stabilization. We show that this method works for both feedforward and recurrent network architectures, trained using either supervised or reinforcement-based learning. This suggests that  employing multiple, complimentary methods, akin to what is believed to occur in the brain, can be a highly effective strategy to support continual learning.
\end{abstract}

% keywords can be removed

\section{Introduction}

Humans and other advanced animals are capable of learning large numbers of tasks during their lifetime, without necessarily forgetting previously learned information. This ability to learn and not forget past knowledge, referred to as continual learning, is a critical requirement to design ANNs that can build upon previous knowledge to solve new tasks. However, when ANNs are trained on several tasks sequentially, they often suffer from "catastrophic forgetting", wherein learning new tasks degrades performance on previously learned tasks. This occurs because learning a new task can alter connection weights away from optimal solutions to previous tasks. 

Given that humans and other animals are capable of continual learning, it makes sense to look toward neuroscientific studies of the brain for possible solutions to catastrophic forgetting.  Within the brain, most excitatory synaptic connections are located on dendritic spines \cite{peters1991fine}, whose morphology shapes the strength of the synaptic connection \cite{kasai2003structure,yuste2001morphological}.  This morphology, and hence the functional connectivity of the associated synapses, can be either dynamic or stable, with lifetimes ranging from seconds to years \cite{kasai2003structure,yoshihara2009dendritic,fischer1998rapid}. Particularly, skill acquisition and retention is associated with the creation and stabilization of dendritic spines \cite{yang2009stably,xu2009rapid}. These results have inspired two recent modelling studies proposing methods that mimic spine stabilization to alleviate catastrophic forgetting \cite{zenke2017continual,kirkpatrick2017overcoming}. Specifically, the authors propose methods to measure the importance of each connection and bias towards solving a task, and then stabilize each according to its importance. Applying these stabilization techniques allows ANNs to learn several ($\leq$10) sequentially trained tasks with only a small loss in accuracy.

However, humans and other animals will likely encounter large numbers (>>100) of different tasks and environments that must be learned and remembered, and it is uncertain whether synaptic stabilization alone can support continual learning across large numbers of tasks. Consistent with this notion, neuroscience studies have proposed that diverse mechanisms operating at the systems \cite{cichon2015branch, tononi2014sleep}, morphological \cite{kasai2003structure,yoshihara2009dendritic}, and transcriptional \cite{kukushkin2017memory} levels all act to stabilize learned information, raising the question as to whether several complementary algorithms are required to support continual learning in ANNs.

In this study, we examine whether another neuroscience-inspired solution, context-dependent gating (XdG), can further support continual learning. In the brain, switching between tasks can disinhibit sparse, highly non-overlapping sets of dendritic branches  \cite{cichon2015branch}. This allows synaptic changes to occur on a small set of dendritic branches for any one task, with minimal interference with synaptic changes that occurred for previous tasks on (mostly) different branches. In this study, we implement a simplified version of this context-dependent gating (XdG) such that sparse and mostly non-overlapping sets of units are active for any one task. The algorithm consists of an additional signal that is unique for each task, and that is projected onto all hidden neurons. Importantly, this algorithm is simple to implement and requires little extra computational overhead.

We tested our method on feedforward networks trained on 100 sequential MNIST permutations \cite{goodfellow2013empirical} and on the ImageNet dataset \cite{imagenet_cvpr09} split into 100 sequential tasks. XdG or synaptic stabilization \cite{zenke2017continual,kirkpatrick2017overcoming}, when used alone, is partially effective at alleviating forgetting across the 100 tasks. However, when XdG is combined with synaptic stabilization, networks can successfully learn all 100 tasks with little forgetting. Furthermore, combining XdG with stabilization allows recurrent neural networks (RNNs), trained using either supervised or reinforcement learning, to sequentially learn 20 tasks commonly used in cognitive and systems neuroscience experiments \cite{yang2017clustering} with high accuracy. Hence, this neuroscience-inspired solution, XdG, when used in tandem with existing stabilization methods, dramatically increases the ability of ANNs to learn large numbers of tasks without forgetting previous knowledge.

\section{Results}

The goal of this study was to develop neuroscience-inspired methods to alleviate catastrophic forgetting in ANNs. Two previous studies have proposed one such method: stabilizing connection weights depending on their importance for solving a task \cite{zenke2017continual,kirkpatrick2017overcoming}. This method, inspired by neuroscience research demonstrating that stabilization of dendritic spines is associated with task learning and retention \cite{yang2009stably,xu2009rapid}, has shown promising results when trained and tested on sequences of $\leq$ 10 tasks. However, it is uncertain how well these methods perform when trained on much larger number of sequential tasks.

We first tested whether these methods can alleviate catastrophic forgetting by measuring performance on 100 sequentially presented digit classification tasks. Specifically, we tested a fully connected feedforward network with two hidden layers (2000 units each, Figure 1A) on the permuted MNIST digit classification task \cite{goodfellow2013empirical}. This involved training the network on the MNIST task for 20 epochs, permuting the 784 pixels in all images with the same permutation, and then training the network on this new set of images. This test is a canonical example of an "input reformatting" problem, in which the input and output semantics (pixel intensities and digit identity, respectively) are identical across all tasks, but the input format (the spatial location of each pixel) changes between tasks \cite{goodfellow2013empirical}.    

We sequentially trained the base ANN on 100 permutations of the image set. Without any synaptic stabilization, this network can classify digits with an accuracy of 98.5\% for any single permutation, but mean classification accuracy falls to 52.5\% after the network is trained on 10 permutations, and to 19.1\% after training on 100 permutations.

\begin{SCfigure}
\includegraphics[width=.51\linewidth]{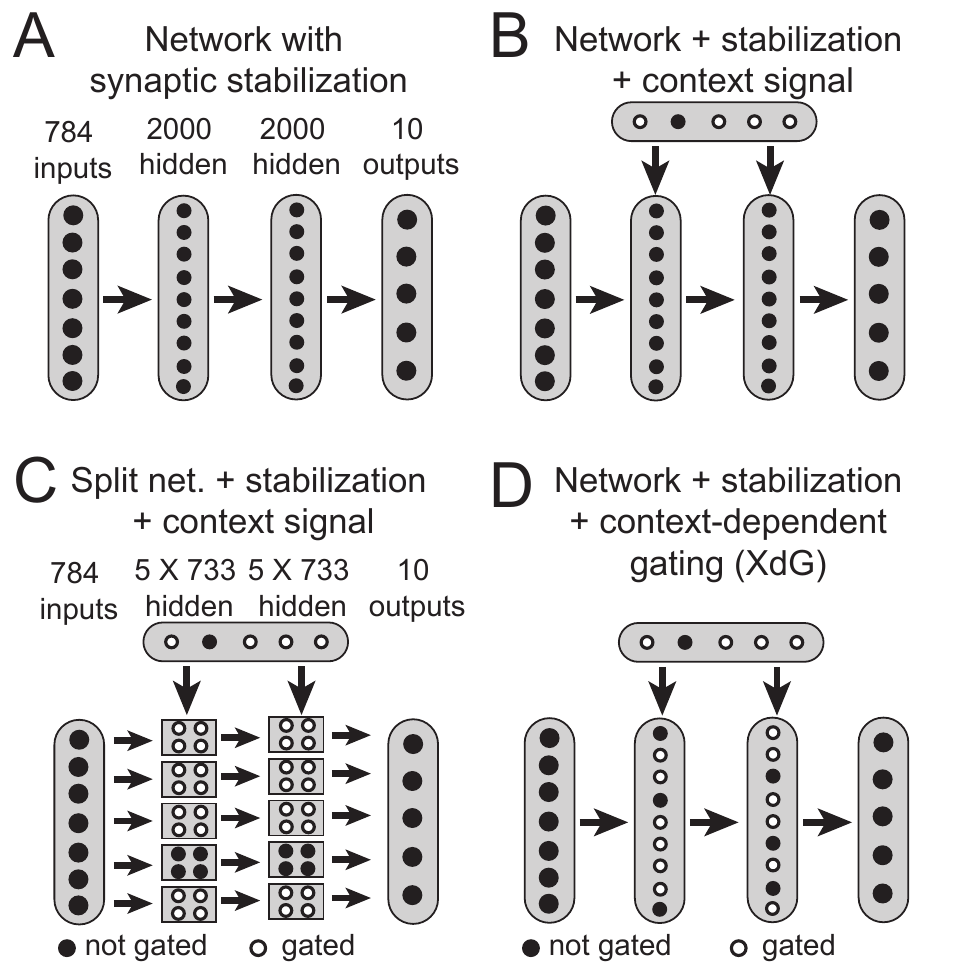}
% \vspace{-0.5cm}
\caption{\small Network architectures for the permuted MNIST task. The ReLu activation function was applied to all hidden units. \textbf{(A)} The baseline network consisted of a multi-layer perceptron with 2 hidden layers consisting of 2000 units each. \textbf{(B)} For some networks, a context signal indicating task identity projected onto the two hidden layers. The weights between the context signal and the hidden layers were trainable. \textbf{(C)} Split networks consisted of 5 independent subnetworks, with no connections between subnetworks. Each subnetwork consisted of 2 hidden layers with 733 units each, such that it contained the same amount of parameters as the full network described in (A). Each subnetwork was trained and tested on 20\% of tasks, implying that for every task, 4 out of the 5 subnetworks was set to zero (fully gated). A context signal, as described in B, projected onto the two hidden layers. \textbf{(D)} Context-dependent gating (XdG) consisted of multiplying the activity of a fixed percentage of hidden units by 0 (gated), while the rest were left unchanged (not gated). The results in Figure 2D involve gating 80\% of hidden units. \vspace{1cm}}
\label{fig:fig1}
\end{SCfigure}

\subsection{Synaptic Stabilization}

Given that this ANN rapidly forgets previously learned permutations of the MNIST task, we next asked how well two previously proposed synaptic stabilization methods, synaptic intelligence (SI) \cite{zenke2017continual} and elastic weight consolidation (EWC) \cite{kirkpatrick2017overcoming}, alleviate catastrophic forgetting. Both methods work by applying a quadratic penalty term on adjusting synaptic values, multiplied by a value quantifying the importance of each connection weight and bias term towards solving previous tasks (see Methods). We note that we use the term "synapse" to refer to both connection weights and bias terms. Briefly, EWC works by approximating how small changes to each parameter affect the network output, calculated after training on each task is completed, while SI works by calculating how the gradient of the loss function correlates with parameter updates, calculated during training. Both stabilization methods significantly alleviate catastrophic forgetting: mean classification accuracies for networks with EWC (green curve, Figure 2A) were 95.3\% and 70.8\% after 10 and 100 permutations, respectively, and mean classification accuracies for networks with SI (magenta curve, Figure 2A) were 97.0\% and 82.3\% after 10 and 100 permutations, respectively. We note that we used the hyperparameters that produced the greatest mean accuracy across all 100 permutations, not just the first 10 permutations (see Methods). Although both stabilization methods successfully mitigated forgetting, mean classification accuracy after 100 permutations was still far below single-task accuracy. This prompts the question of whether an additional, complementary method can raise performance even further.

\subsection{Context Signal}
One possible reason why classification accuracy decreased after many permutations was that ANNs were not informed as to what permutation, or context, was currently being tested. In contrast, context-dependent signals in the brain, likely originating from areas such as the prefrontal cortex, project to various cortical areas and allow neural circuits to process incoming information in a task-dependent manner \cite{engel2001dynamic,johnston2007top,miller2001integrative}. Thus, we tested whether such a context signal improves mean classification accuracy. Specifically, a one-hot vector (N-dimensional consisting of N-1 zeros and 1 one), indicating task identity, projected onto the two hidden layers. The weights projecting the context signal could be trained by the network.

We found that networks including parameter stabilization combined with a context signal had greater mean classification accuracy (context signal with SI = 89.6\%, with EWC = 87.3\%, Figure 2B) than synaptic stabilization alone. However, the mean classification accuracy after 100 tasks still falls short of single-task accuracy, suggesting that contextual information alone is insufficient to alleviate forgetting.

\subsection{Split Network}

Recent neuroscience studies have highlighted how context-dependent signaling not only allows various cortical areas to process information in a context-dependent manner, but selectively inhibits large parts of the network \cite{kuchibhotla2017parallel,otazu2009engaging}. This inhibition potentially alleviates catastrophic forgetting, provided that changes in synaptic weights only occur if their pre- and post-synaptic partners are active during the task, and are frozen otherwise.

To test this possibility, we split the network into 5 subnetworks of equal size (Figure 2C). Each subnetwork contained 733 neurons in each hidden layer, so that the number of connection weights in this split network matched the number of free parameters in the full network. For each permutation, one subnetwork was activated, and the other four were fully inhibited. Furthermore, the context signal described in Figure 2B projected onto the hidden layers. This architecture achieved greater mean classification accuracies (split networks with context signal and SI = 93.1\%, and EWC = 91.4\%) than full networks with stabilization combined with a context signal.

\begin{figure}
\centering
\includegraphics[width=.7\linewidth]{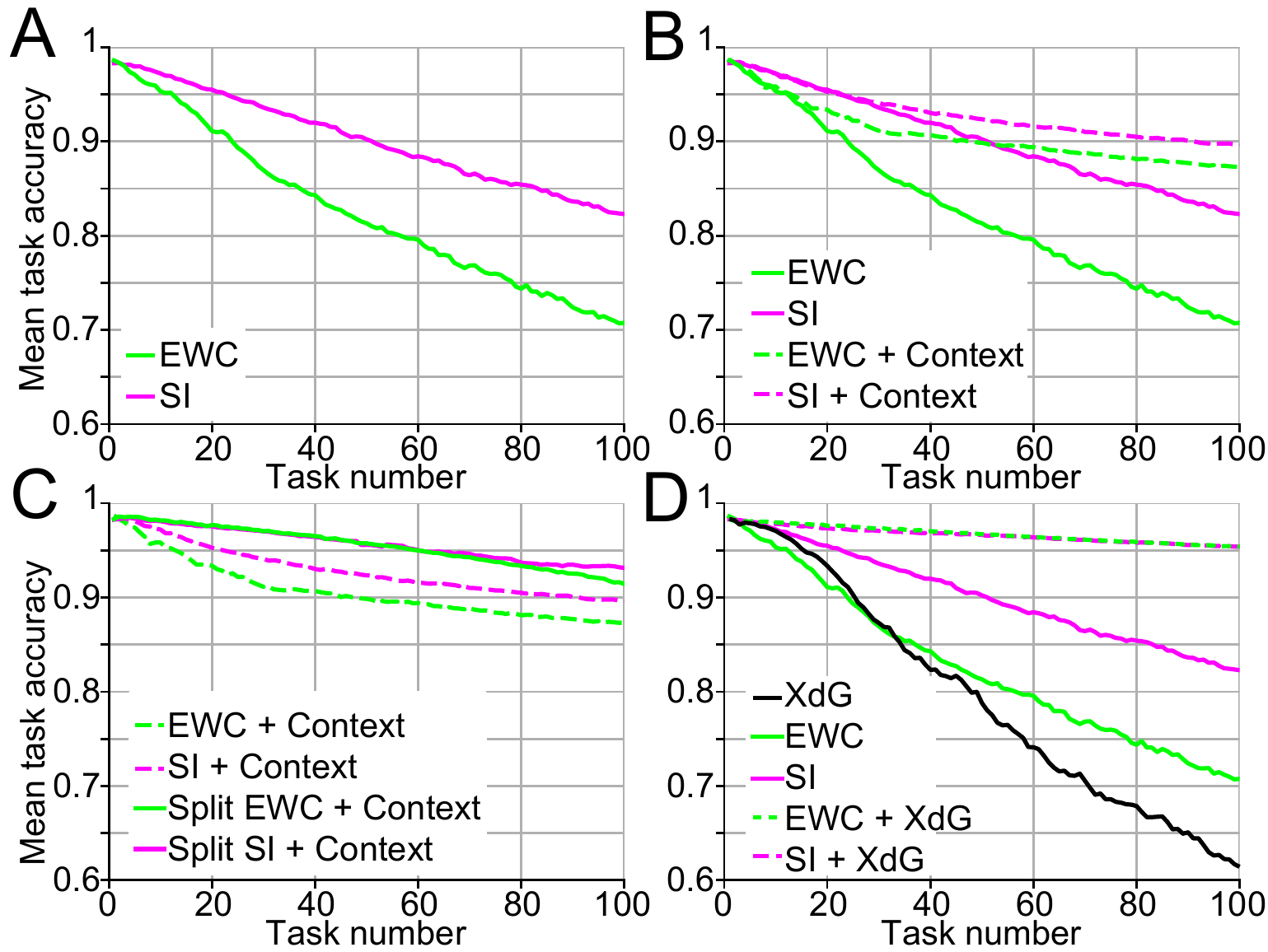}
\caption{\small Task accuracy on the permuted MNIST benchmark task. All curves show the mean classification accuracy as a function of the number of tasks the network was trained on, where each task corresponds to a different random permutation of the input pixels. \textbf{(A)} The green and magenta curves represent the mean accuracy for networks with EWC and with SI, respectively. \textbf{(B)} The solid green and magenta curves represent the mean accuracy for networks with EWC and with SI, respectively (same as in A), and the dashed green and dashed magenta curves represent the mean accuracy for networks with a context signal combined with EWC or SI, respectively. \textbf{(C)} The dashed green and magenta curves represent the mean accuracy for networks with a context signal combined with EWC or SI, respectively (same as in B), and the solid green and magenta curves represent the mean accuracy for split networks with a context signal combined with EWC or SI, respectively. \textbf{(D)} The solid green and magenta curves represent the mean accuracy for networks with EWC or SI, respectively (same as in A), the black curve represents the mean accuracy of networks with XdG used alone, and the dashed green and magenta curves represent the mean accuracy for networks with XdG combined with EWC or SI, respectively.}
\vspace{2mm}
\hrule
\vspace{-4mm}
\label{fig:fig2}
\end{figure}

\subsection{Context-Dependent Gating}

These results suggests that context-dependent inhibition can potentially allow networks to learn sequentially presented tasks with less forgetting. However, splitting networks into a fixed number of subnetworks a priori may be impractical for more real-world tasks. Furthermore, this method forces each subnetwork to learn multiple tasks, whereas greater classification accuracies might be possible if unique, partially-overlapping sets of synapses are responsible for learning each new task. Thus, we tested a final method, context-dependent gating (XdG), in which the activity of $X\%$ of hidden units, randomly chosen, was multiplied by 0 (gated), while the activity of the other $(1-X)\%$ was left unchanged. In this study, we gated 80\% of hidden units per task. The identity of the gated units was fixed during training and testing for the specific permutation, and a different set of fully gated hidden units was chosen for each permutation. When XdG was used alone (black curve, Figure 2D), mean accuracy was 97.1\% after 10 tasks, and 61.4\% across all 100 permutations. However, when XdG was combined with SI or EWC, mean classification accuracy was 95.4\% for both stabilization methods (dashed green and magenta lines), greater than any of the previous methods we tested. Thus, while XdG alone does not support continual learning, it becomes highly effective when paired with existing synaptic stabilization methods.

The optimal percentage of units to gate is a compromise between keeping a greater number of units active, increasing the network's representational capacity, and keeping a greater numbers of units gated, which decreases the number of connection weight changes and forgetting between tasks. For the permuted MNIST task, we found that mean classification accuracy peaked when between 80\% or 86.7\% of units were gated (values of 95.4\% and 95.5\%, respectively) (Figure S1). We note that this optimal value depends on the network size and architecture, and the number of tasks upon which the network is trained.

XdG combined with synaptic stabilization allowed networks to learn 100 sequentially trained tasks with minimal loss in performance, with accuracy dropping from 98.2\% on the first task to a mean of 95.4\% across all 100 tasks. This result raises the question of whether XdG allows networks to learn even more tasks with only a gradual loss in accuracy, or if they would instead reach a critical point where accuracy drops abruptly. Thus, we repeated our analysis for 500 sequentially trained tasks, and found that XdG combined with stabilization allowed for continual learning with only a gradual loss of accuracy over 500 tasks (XdG combined with  SI = 90.7\%, Figure S2). In comparison, mean accuracy for stabilization alone decreased more severely (SI = 54.9\%).

\begin{SCfigure}
\includegraphics[width=.5\linewidth]{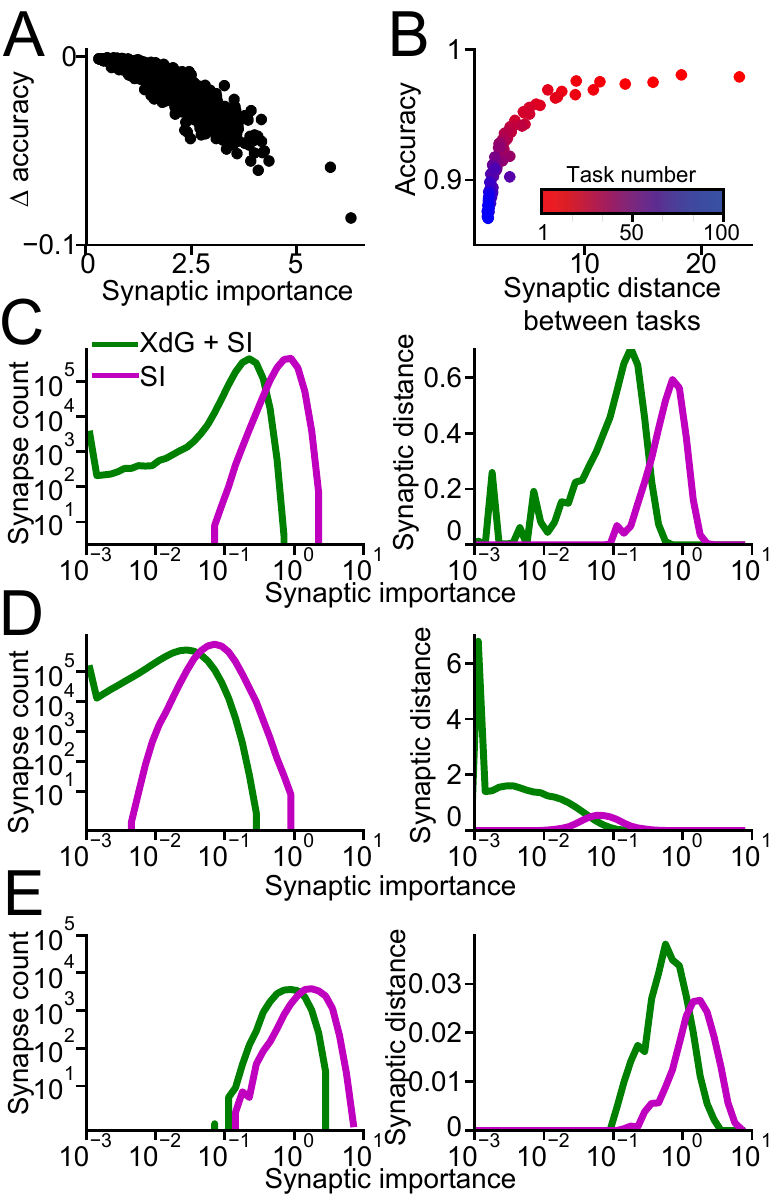}
\caption{\small Analyzing the interaction between XdG and synaptic stabilization. \textbf{(A)} The effect of perturbing synapses of various importance is shown for a network with SI that was sequentially trained on 100 MNIST permutations. Each dot represents the change in mean accuracy (y-axis) after perturbing a single synapse, whose importance is indicated on the x-axis. For visual clarity, we show the results from 1000 randomly selected synapses chosen from the connection weights to the output layer.  \textbf{(B)} Scatter plot showing the Euclidean distance in synaptic values measured measured before and after training on each MNIST permutation (x-axis) versus the accuracy the network achieved on each new permutation. The task number in the sequence of 100 MNIST permutation is indicated by the red to blue color progression. \textbf{(C)} Left panel: histogram of synaptic importances from the connections between the input layer and the first hidden layer (layer 1), for networks with XdG (green curve) and without (magenta curve). Right panel: Synaptic distance, measured before and after training on the 100th MNIST permutation, for groups of synapses binned by importance.  \textbf{(D)} Same as (C), except for the synapses connecting the first and second hidden layers (layer 2). \textbf{(E)} Same as (C), except for the synapses connecting the second hidden layer and the output layer (layer 3). \vspace{2cm}}
\label{fig:fig3}
\end{SCfigure}

\subsection{Analyzing the interaction between XdG and synaptic stabilization}

We would like to understand why XdG combined with synaptic stabilization better alleviates forgetting compared to stabilization alone. We hypothesize that to accurately learn many sequential tasks with little forgetting, the network must balance two competing demands: it must stabilize synapses that are deemed important for previous tasks, yet remain flexible so that it can adjust synaptic values by a sufficient amount to accurately learn new tasks.  To demonstrate the first point, which is the basis for SI \cite{zenke2017continual} and EWC \cite{kirkpatrick2017overcoming}, we trained a network with stabilization (SI) on 100 MNIST permutations, and then measured the mean accuracy across all permutations after perturbing individual synaptic values (see Methods for details regarding analysis). As expected, perturbing more important synapses degraded accuracy more than perturbing less important synapses (R = -0.904, Figure 3A).

To demonstrate that learning new tasks requires flexible synaptic values that can be adjusted sufficient amounts to learn new tasks, we show the network's accuracy on each new MNIST permutation it learns (y-axis, Figure 3B) versus the distance between the synaptic values measured before and after training on each MNIST permutation (x-axis). For SI and EWC, synaptic importance values accumulates across tasks (see Methods), leading to greater stabilization and less flexibility as more tasks are learned. For the first several tasks (red dots), synapses require less stabilization, allowing the network to adjust synapse values by relatively large amounts to accurately learn each new task. However, as the network learns increasing numbers of tasks (blue dots), synaptic importances, and hence stabilization, increases, preventing the network from adjusting synapse values large amounts between tasks, decreasing accuracy on each new task.

To help us understand why XdG combined with stabilization better satisfies this trade-off compared to stabilization alone, in the left panels of Figures 3C-E we show the distribution of importance values for synapses connecting the input and first hidden layers (referred to as layer 1, panel C), connecting the first and second hidden layers (layer 2, panel D), and connection the second hidden and output layers (layer 3). For all three layers, the mean importance values were lower for networks with XdG (layer 1, panel C: SI = 0.793, SI + XdG = 0.216; layer 2, panel D, SI = 0.076, SI + XdG = 0.026; layer 3, panel D, SI = 1.81, SI + XdG = 0.897). We hypothesized that having larger number of synapses with low importance is beneficial, as the network could adjust those synapses to learn new tasks with minimal disruption to performance on previous tasks. 

To confirm this, we binned the synapses by their importance, and calculated the Euclidean distance in synaptic values measured before and after training on the 100th MNIST permutation (right panels, Figures 3C-E). These panels suggest that synaptic distances are greater for networks with XdG combined with SI, and that larger changes in synaptic values are more confined to synapses with low importances. Thus, these panels suggest that networks with XdG makes large adjustments to synapses of relatively little importance, primarily in layers 1 and 2.

To confirm, we calculated the synaptic distance calculated using all synapses from each layer, and found that it was greater for networks with XdG combined with SI (layer 1, panel C, SI = 1.21, SI + XdG = 1.65; layer 2, panel D, SI = 1.26, SI + XdG = 8.54; layer 3, panel E, SI = 0.06, SI + XdG = 0.09). Furthermore, the distance for each synaptic value multiplied by its importance was lower for networks with XdG combined with SI (layer 1, panel C, SI = 457.33, SI + XdG = 83.56; layer 2, panel D, SI = 109.75, SI + XdG = 17.83; layer 3, panel E, SI = 12.15, SI + XdG = 3.16). Thus, networks with XdG combined with stabilization can make larger adjustments to synapse values to accurately learn new tasks, and simultaneously make smaller adjustments to synapses with high importance.

By gating 80\% of hidden units, 96\% of the weights connecting the two hidden layers and 80\% of all other parameters are not used for any one task. This allows networks with XdG to maintain a reservoir of synapses that have not been previously used, or used sparingly, that can be adjusted by large amounts to learn new tasks without disrupting performance on previous tasks.

\subsection{XdG on the ImageNet Dataset}

A simplifying feature of the permuted MNIST task is that the input and output semantics are identical between tasks. For example, output unit 1 is always associated with the digit 1, output unit 2 is always associated with the digit 2, etc. We wanted to ensure that our method generalizes to cases in which the output semantics are similar between tasks, but not identical. Thus, we tested our method on the ImageNet dataset, which comprises approximately 1.3M images, with 1000 class labels. For computational efficiency, we used images that were downscaled to $32 \times 32$ pixels from the more traditional resolution of $256 \times 256$.  We divided the dataset into 100 sequential tasks, in which the first 10 labels of the ImageNet dataset were assigned to task 1, the next 10 labels were assigned to task 2, etc. The 10 class labels used in each of the 100 tasks are shown in Table S1. Such a test can be considered an example of a "similar task" problem as defined by \cite{goodfellow2013empirical}.

We tested our model on the ImageNet tasks using two different output layer architectures. The first was a "multi-head" output layer, which consisted of 1000 units, one for each image class. The output activity of 990 output neurons not applicable to the current task was set to zero.

To measure the maximum obtainable accuracy our networks could achieve when sequentially trained on the 100 tasks, we trained and tested networks without stabilization and with resetting synaptic values between tasks, and measured their accuracy at learning each new task. We disregarded accuracy on previous tasks. These networks achieved a mean accuracy of 56.5\%, representing the maximum any network of this architecture could achieve across the sequence of 100 tasks.  In comparison, mean accuracy across 100 sequentially trained tasks using the "multi-head" output layer was 36.7\% without synaptic stabilization, and adding synaptic stabilization almost fully alleviated forgetting (SI = 51.1\%, EWC = 54.9\%; Figure 4A)

The multi-headed network architecture, while potentially effective, can be impractical for real-world implementations as it requires one output neuron for each possible output class that the network might encounter, which might not be known \textit{a priori}. Thus, we also tested a "single-head" output layer, which consisted of only 10 units, and the activity of these 10 units was associated with different image classes in different tasks. Mean classification accuracy for this more challenging architecture was substantially lower (without stabilization = 10.5\%, SI = 12.3\%, EWC = 11.6\%; Figure 4A). 

We wanted to know whether our method could alleviate forgetting for this more challenging architecture, in which output units were associated with different image classes. Thus, we repeated our analysis used for Figures 2B-D. We found that adding a context signal substantially increased mean accuracy (context signal with SI = 42.7\%, with EWC = 44.4\%, Figure 4B). Splitting networks into five subnetworks did not lead to any improvement over using stabilization and a context signal alone (split networks with context signal and SI = 40.0\%, and EWC = 42.5\%, Figure 4C). Next, we trained networks with XdG, in which we  gated 80\% of hidden units. Mean accuracy for XdG alone (black curve, Figure 4D) was greater than SI or EWC used alone, but was less than networks with stabilization and a context signal, or split networks (XdG = 28.1\%). However, when XdG was combined with SI or EWC, networks could learn all 100 tasks with little forgetting (XdG with SI = 50.7\%, with EWC = 52.4\%, Figure 4D). Thus, XdG used in tandem with synaptic stabilization can alleviate catastrophic forgetting even when the output domain differs between tasks.

\begin{figure}%[tbhp]
\centering
\includegraphics[width=.7\linewidth]{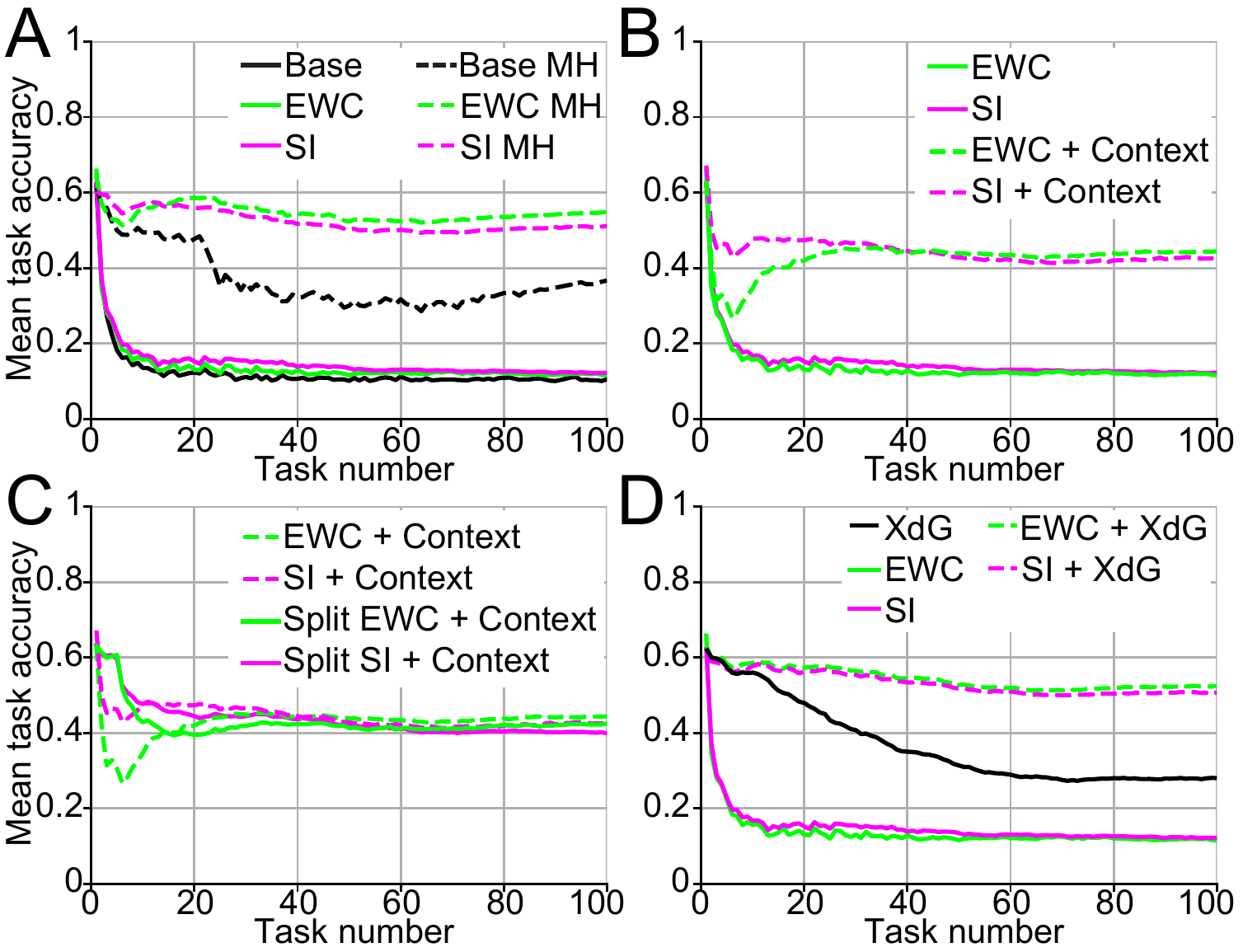}
\caption{\small Similar to Figure 2, except showing the mean image classification accuracy for the ImageNet dataset split into 100 sequentially trained sections. \textbf{(A)} The dashed black, green and magenta curves represent the mean accuracies for multi-head networks without synaptic stabilization, with EWC or SI, respectively. The solid black, green and magenta curves represent the mean accuracies for non multi-head networks without synaptic stabilization, with EWC or SI, respectively. All further results involve non multi-head networks. \textbf{(B)} The solid green and magenta curves represent the mean accuracies for networks with EWC or SI, respectively (same as in A). The dashed green and magenta curves represent the mean accuracies for networks with a context signal combined with EWC or SI, respectively. \textbf{(C)} The dashed green and magenta curves represent the mean accuracies for networks with a context signal combined with EWC or SI, respectively (same as in B). The solid green and magenta curves represent the mean accuracies for split networks, with a context signal combined with EWC or SI, respectively. \textbf{(D)} The black curve represents the mean accuracy for networks with XdG used alone. The solid green and magenta curves represent the mean accuracies for networks with EWC or SI, respectively (same as in A). The dashed green and magenta curves represent the mean accuracies for networks with XdG combined with EWC or SI, respectively.}
\vspace{2mm}
\hrule
\vspace{-4mm}
\label{fig:fig4}
\end{figure}

\subsection{XdG on Recurrent Neural Networks}

The sequential permuted MNIST and ImageNet tests demonstrated that XdG, combined with synaptic stabilization, can alleviate forgetting for feedforward networks performing classification tasks, trained using supervised learning. We next demonstrate that our method generalizes and performs well in other task conditions. Thus, we trained recurrent neural networks (RNNs) on 20 sequentially presented cognitively demanding tasks \cite{yang2017clustering}. These 20 tasks are similar to those commonly used in neuroscience experiments, and involve decision-making, working memory, categorization and inhibitory control. In a physical setting, all tasks involve one or multiple visual motion stimuli, plus a fixation cue, that are presented to a subject, who reports their decisions by either maintaining gaze fixation (akin to withholding its response) or by performing a saccadic eye movement to one of several possible target locations.

\begin{SCfigure}%[tbhp]
\includegraphics[width=.6\linewidth]{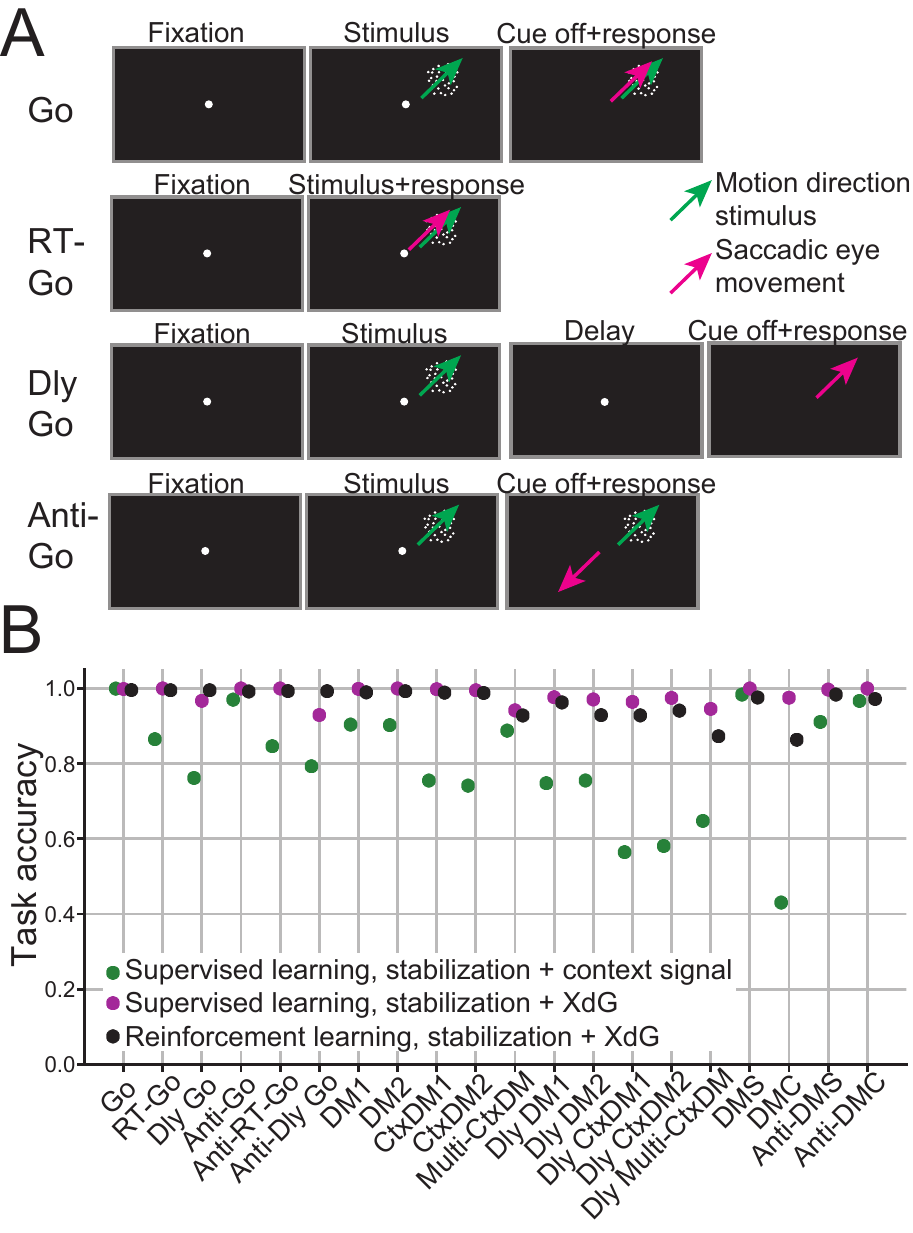}
\caption{\small Task accuracy of recurrent networks sequentially trained on 20 cognitive-based tasks. \textbf{(A)} Schematics of the first four tasks. All trials involve a motion direction stimulus (represented by the white dot pattern and green motion direction arrow), a fixation cue (represented by the white centrally located dot), and a action response using an eye saccade (represented by a magenta arrow). \textbf{(B)} The mean accuracy across all 20 tasks for the three network configurations. Green dots represent mean accuracy for networks with stabilization (SI) combined with a rule cue, trained using supervised learning. Magenta dots represent the mean accuracy for networks with stabilization (SI) combined with XdG, trained using supervised learning. Black dots represent the mean accuracy for networks with stabilization (SI) combined with XdG, trained using reinforcement learning. \vspace{3cm}}
\label{fig:fig5}
\end{SCfigure}

Schematics of the first 4 tasks are shown in Figure 5A. In the Go task, trials begin with a fixation period lasting a random duration, followed by a motion direction stimulus (represented by a white dot pattern and white arrow) that could occur in one of two locations. As soon as the fixation cue (white centrally located dot) disappears, the network should respond by moving in the dire the motion stimulus (represented by the red arrow). The RT-Go task is similar to the task above, except crucially, the network should ignore the fixation cue and respond as soon as it is presented with the motion stimulus. In the Delay Go task, the motion stimulus is briefly presented, and the network must maintain the stimulus direction in short-term memory across a delay period until the fixation cue disappears, at which time it can respond. Lastly, the Anti-Go is similar to the Go task, except that the network must respond in the direction 180 degrees opposite to the motion direction stimulus. Thus, to learn many tasks, the network must learn to ignore, or to actively work against, information it has learned in previous tasks. Full description of all 20 tasks is provided in the Supplemental Information.

Our RNN consisted of 256 LSTM cells \cite{hochreiter1997long} that received input from 64 motion direction tuned input units and 4 fixation tuned units. It projected onto nine output units; one unit represented the choice to maintain fixation (i.e. withhold a response), and the other eight represented responses in eight different directions.

To assess how various methods can learn these tasks without forgetting, we trained networks on all tasks sequentially, and then measured accuracy for each task. We first trained RNNs using standard supervised learning, in which the network parameters were adjusted to minimize the difference between the actual network output and the target output. Using this approach, RNNs equipped with synaptic stabilization (SI) and a context-signal (green dots, Figure 5B) achieved a mean task accuracy of  80.0\%, with a range of 43.1\% to 100.0\%. In comparison, networks with stabilization combined with XdG achieved a mean accuracy  98.2\%, with a range of 92.9\% to 100.0\%.  

The networks described above were all trained using supervised learning. If our method is to work in more practical settings, and if something akin to this method is implemented in the brain, then it should be compatible with reinforcement-learning based training. Thus, we repeated our test on networks with synaptic stabilization (using a modified version of SI compatible with reinforcement learning, see Methods) combined with XdG. Mean accuracy across all 20 tasks was 96.4\%, with a range of 86.4\% to 100.0\%. Thus, these results demonstrate that XdG, when combined with synaptic stabilization, can allow RNNs to learn a large sequence of cognitively demanding tasks, and is compatible with supervised or reinforcement-learning.

\section{Discussion}

In this study, we have shown that context-dependent gating (XdG), used in conjunction with previous methods to stabilize synapses, can alleviate catastrophic forgetting in feedforward and recurrent networks trained using either supervised or reinforcement learning, on large numbers of sequentially presented tasks. This method is simple to implement and computationally inexpensive. Importantly, this study highlights the effectiveness of employing multiple, complimentary strategies to alleviate catastrophic forgetting, as opposed to relying on a single strategy. Such an approach would appear to be consistent with that used by the brain, which employs a diverse set of mechanisms to combat forgetting and promote consolidation \cite{cichon2015branch,yang2009stably,tononi2014sleep,yoshihara2009dendritic}.

\subsection{Transfer learning}
Humans and other advanced animals can rapidly learn new rules or tasks, in contrast to the thousands of data points usually required for ANNs to accurately perform new tasks. This is because most new tasks are at least partially similar to tasks or contexts that have been previously encountered, and one can use past knowledge learned from these similar tasks to help learn the rules and contingencies of the new task. This process of using past knowledge to rapidly solve new tasks is referred to as transfer learning, and several groups have recently proposed how ANNs can implement this form of rapid learning \cite{santoro2016one,lake2013one,fernando2017pathnet}. 

Although the XdG method we have proposed in this study likely does not support transfer learning in its current form, one could speculate how such a signal could be modified to perform this function. Suppose that specific ensembles of units underlie the various cognitive processes, or building blocks, required to solve different tasks. Then learning a new task would not require relearning entirely new sets of connection weights, but rather implementing a context-dependent signal that activates the necessary building blocks, and facilitates their interaction, in order to solve the new task. PathNet \cite{fernando2017pathnet}, a recent method in which a genetic algorithm selects a subset of the network to utilize for each task, is one prominent example of how selective gating can facilitate transfer. Although it is computationally intensive, requires freezing previously learned synapses, and has only demonstrated transfer between two sequential tasks, it clearly shows that gating specific network modules can allow the agent to reuse previously learned information, decreasing the time required to learn a new task. 

We believe that further progress towards transfer learning will require progress along several fronts. First, as humans and other advanced animals can seamlessly switch between contexts, novel algorithms that can rapidly identify network modules applicable to the current task are needed. Second, algorithms that can identify the current task or context, compare it to previously learned contexts, and then perform the required gating based on this comparison, are also required. The method proposed in this study clearly lacks this capability, and developing this ability is crucial if transfer learning is to be implemented in real-world scenarios. 

Third, and perhaps most important, is that the network must represent learned information in a format to support the above two points. If learned information is located diffusely throughout the the network, then activating the relevant circuits and facilitating their interaction might be impractical. Strategies that encourage the development of a modular representation, as is believed to occur in brain \cite{bassett2011dynamic}, might be required to fully implement continual and transfer learning \cite{velez2017diffusion}. We believe that \textit{in vivo} studies examining how various cortical and subcortical areas underlie task learning will be provide invaluable data to guide the design of novel algorithms that allow ANNs to rapidly learn new information in a wide range of contexts and environments. 

\subsection{Related Methods to Alleviate Catastrophic Forgetting}

The last several years has seen the development of several methods to alleviate catastrophic forgetting in neural networks. Earlier approaches, such as Progressive Neural Networks \cite{rusu2016progressive} and Learning Without Forgetting \cite{li2017learning}, achieved success by adding additional modules for each new task. Both methods include the use of a multi-head output layer (see Figure 4A), and while effective, in this study we were primarily interested in the more general case in which the network size cannot be augmented to support each additional task, and in which the same output units must be shared between tasks.

Other studies are similar in spirit to SI \cite{zenke2017continual} and EWC \cite{kirkpatrick2017overcoming} in that they propose methods to stabilize important network weights and biases, \cite{aljundi2017memory,nguyen2017variational}, or to stabilize the linear space of parameters deemed important for solving previous tasks \cite{he2017overcoming}. While we did not test the performance of these methods, we note that like EWC and SI, these algorithms could in theory also be combined with XdG to potentially further mitigate catastrophic forgetting.

Another class of studies have proposed similar methods that also gate parts of the network. Aside from PathNet \cite{fernando2017pathnet} described above, recent methods have proposed gating network connection weights \cite{mallya2017packnet} or units \cite{serra2018overcoming} to alleviate forgetting. The two key differences between our method and theirs are: 1) gating is defined \textit{a priori} in our study, which increases computational efficiency, but potentially makes it less powerful, and 2) we propose to combine gating with parameter stabilization, while their methods involve gating alone. We tested the Hard Attention to Task (HAT) method \cite{serra2018overcoming} and found that it could learn 100 sequential MNIST permutations with a mean accuracy of 93.0\% (Figure S3), greater than networks with EWC or SI alone. That said, combining SI and XdG still outperforms HAT (95.8\% when using the same number of training epochs as HAT, Figure S3), and is computationally less demanding. This further highlights the advantage of using complementary methods to alleviate forgetting, but also suggests that adding synaptic stabilization to HAT could allow it to further mitigate forgetting.

\subsection{Summary}

Drawing inspiration from neuroscience, we propose that context-dependent gating, a simple to implement method with little computational overhead, can allow feedforward and recurrent networks, trained using supervised or reinforcement learning, to learn large numbers of tasks with little forgetting when used in conjunction with synaptic stabilization. Future work will build upon this method so that it not only alleviate catastrophic forgetting, but can also support transfer learning between tasks. 

\section{Methods}

Network training and testing was performed using the machine-learning framework TensorFlow \cite{abadi2016tensorflow}. All code is available at https://github.com/nmasse/Context-Dependent-Gating.

\subsection{Feedforward Network Architecture}

For the MNIST task, we used a fully connected network consisting of 784 input units, two hidden layers of 2000 hidden units each, and 10 outputs. We did not use a multi-head output layer, and thus the same 10 output units were used for all permutations. The ReLU activation function and Dropout \cite{srivastava2014dropout} with a 50\% drop percentage was applied to all hidden units. The softmax nonlinearity was applied to the units in the output layer.

The ImageNet network included four convolutional layers. The first two layers used 32 filters with $3 \times 3$ kernel size and $1 \times 1$ stride, and the second two layers used 64 filters with the same kernel size and stride.  Max pooling with a $2 \times 2$ stride was applied after layers two and four, along with a 25\% drop rate.  Gating was not applied to the convolutional layers. After the four convolutional layers were two full connected hidden layers of 2000 units each. As above, the ReLu activation function and a 50\% drop rate was applied to the hidden units in the two fully connected layers. The softmax activation function was applied to the output layer. We primarily used a single-head output layer, with 10 units in the output layer that were used in all tasks (Figures 4A-D). For comparison, we also tested a multi-head output layer (Figure 4A only) consisting of 1000 output units wherein 990 of the units were masked for any one task.

For computational efficiency, we first trained the ImageNet network on a different, yet similar, dataset, which combined the 50000 images of the CIFAR-10 dataset with the 50000 images of the CIFAR-100 dataset. After training, we fixed the parameters in the convolutional layers, and then trained the parameters in the fully-connected (and not the convolutional) layers of the network on the 100 tasks of the ImageNet dataset.

\subsection{Recurrent Network Architecture}

All RNNs consisted of 256 LSTM cells \cite{hochreiter1997long} that projected onto 1 unit representing the choice to maintain fixation (i.e. withhold a response), and 8 units representing responses towards eight different directions. The softmax nonlinearity was applied to activity in the output layer. Input into the RNN consisted of 4 fixation tuned units and 2 sets of 32 motion direction tuned units, in which each set represents visual motion input from one of two spatial locations. For RNNs that used a context signal (green dots, Figure 5B), the input consisted of an additional one-hot vector of length 20 representing the current task.

\subsection{Network Training and Testing}

For all networks, parameters were trained using the Adam optimizer \cite{kingma2014adam} ($\eta = 0.001$, $\beta_1 = 0.9$, $\beta_2 = 0.999$). The optimizer state was reset between tasks. 

When training the feedforward and recurrent networks using supervised learning, we used the cross-entropy loss function. We trained the networks for 20 epochs for each MNIST task, 40 epochs for each ImageNet task, and for 6000 training batches on each cognitive based task.
We used a batch size of 256 and a learning rate of 0.001.  We tested classification accuracy on each task using 10 batches; for the MNIST and ImageNet tasks, the test images were kept separate from the training images.

In addition to supervised learning, we also trained RNNs using the actor-critic reinforcement learning method \cite{barto1983neuronlike},

\begin{equation}
\mathcal{R}_{\tau} = \sum_{t=\tau}^{T} \gamma^{t-\tau} r_{t},
\end{equation}

where $\gamma \in [0,1)$ is the discount factor and $r_{t}$ is the reward given at time $t$. For this method, the RNN output consists of a nine-dimensional policy vector that maps the activity of the RNN into a probability distribution over actions, and a value scalar which estimates the future discounted reward. Specifically, we denote the policy output as $\pi_{\theta}(a_{t}|h_{t})$, where $\theta$ represents the network parameters, $a_{t}$ the vector of possible actions at time $t$, and $h_{t}$ as the output activity of the LSTM cells. We also denote the value output as $V_{\theta}(h_{t})$. The loss function can be broken down into expressions related to the value output, the policy output, and an entropy term that encourages exploration. First, the network should minimize the mean squared error between the predicted and expected discounted future reward,

\begin{equation}
\mathcal{L}_{V} = \frac{1}{2T} \sum_{t=1}^{T}[V_{\theta}(h_{t}) - r_{t} - \gamma V_{\theta}(h_{t+1})]^{2}.
\end{equation}

Second, the network should maximize the logarithm of choosing actions with large advantage values (defined below)

\begin{equation}
\mathcal{L}_{P} = - \frac{1}{T} \sum_{t=1}^{T} A_{t} \log(\pi_{\theta}(a_{t}|h_{t})),
\end{equation}

In this study, we use the generalized advantage estimation \cite{schulman2015high}, which represents the difference between the expected and the actual reward:

\begin{equation}
A_{t} = r_{t} + \gamma V_{\theta}(h_{t+1}) - V_{\theta}(h_{t})
\end{equation}

We note that when calculating the gradient of the policy loss function, one treats the advantage function as a fixed value (i.e. one does not compute the gradient of the advantage function).
 
We also include an entropy term that the network should maximize that encourages exploration by penalizing overly confident actions:

\begin{equation}
\mathcal{L}_{H} = - \frac{1}{T} \sum_{t=1}^{T} \pi_{\theta}(a_{t}|h_{t}) \log(\pi_{\theta}(a_{t}|h_{t})).
\end{equation}

We obtain the overall loss function by combining all three terms:

\begin{equation}
\mathcal{L} =\mathcal{L}_{P} + \beta \mathcal{L}_{V} - \alpha \mathcal{L}_{H},
\end{equation}

where $\alpha$ and $\beta$ control how strongly the entropy and value loss functions, respectively, determine the gradient.

\subsection{Synaptic Stabilization}

The XdG method proposed in this study was used in conjunction with one of two previously proposed methods to stabilize synapses: synaptic intelligence (SI) \cite{zenke2017continual} and elastic weight consolidation (EWC) \cite{kirkpatrick2017overcoming}. Both methods work by adding a quadratic penalty term to the loss function that penalizes weight changes away from their values before starting training on a new task:

\begin{equation}
\mathcal{L} = \mathcal{L}_{k} + c \sum_{i} \Omega_i (\theta_i - \theta^{prev}_{i})^2,
\end{equation}

where $\mathcal{L}_{k}$ is the loss function of the current task $k$, $c$ is a hyperparameter that scales the strength of the synaptic stabilization, $\Omega_i$ represents the importance of each parameter $\theta_i$ towards solving the previous tasks, and $\theta^{prev}_{i}$ is the parameter value at the end of the previous task. 

For EWC, $\Omega^{k}_i$ is calculated for each task $k$ as the diagonal elements of the Fisher information $\mathcal{F}$:

\begin{equation}
\mathcal{F} = \mathbb{E}_{x \sim \mathcal{D}^{k}, y \sim p_{\theta}(y | x)}\Bigg[\Big(\frac{\partial \log{p_{\theta}(y | x)}}{\partial \theta}\Big)\Big(\frac{\partial \log{p_{\theta}(y | x)}}{\partial \theta}\Big)^{T}\Bigg], 
\end{equation}

where the inputs $x$ are sampled from the data distribution for task $k$, $\mathcal{D}^{k}$, and the labels $y$ are sampled from the model $p_{\theta}$. We calculated the Fisher information using an additional 32 batches of 256 training data points after training on each task was completed.

For SI, for each task $k$, we first calculated the product between the change in parameter values, $\theta(t)-\theta(t-1)$, with the negative of the gradient of the loss function $\frac{\partial \mathcal{L}_{k}(t)}{\partial \theta}$, summed across all training batches $t$:

\begin{equation}
\omega^{k}_{i} = \sum_{t} (\theta_{i}(t)-\theta_{i}(t-1)) \frac{-\partial \mathcal{L}_{k}(t)}{\partial \theta_{i}}
\end{equation}

We then normalize this term by the total change in each parameter $\Delta \theta_{i} = \sum_{t} (\theta_{i}(t)-\theta_{i}(t-1))$ plus a damping term $\zeta$:

\begin{equation}
\Omega^{k}_{i} = \max \Bigg(0, \frac{\omega^{k}_{i}}{(\Delta \theta_{i})^{2} + \zeta} \Bigg).
\end{equation}

Finally, for both EWC and SI, the importance of each parameter $i$ at the start of task $m$ is the sum of $\Omega^{k}_{i}$ across all completed tasks:

\begin{equation}
\label{eq:Om}
\Omega_{i} = \sum_{k<m} \Omega^{k}_{i}.
\end{equation}

\subsection{Parameter Stabilization for Reinforcement Learning}

The method described above to calculate SI is ill-suited for reinforcement learning, as the policy loss function depends on the estimated discounted future reward, which can be inaccurate, especially early in training. Thus, we first calculate the product between the change in parameter values, $\theta(t)-\theta(t-1)$, and the change in the mean reward obtained for each batch, $R(t)-R(t-1)$, summed across all training batches $t$:

\begin{equation}
\omega^{k}_{i} = \sum_{t} (\theta_{i}(t)-\theta_{i}(t-1)) (R(t) - R(t-1)).
\end{equation}

To normalize, we first calculate the sum of the absolute value of this product that represents the maximum value the above equation can obtain if sign of the change in parameter and the sign of the change in mean reward are always aligned: 

\begin{equation}
\overline{\omega}^{k}_{i} = \sum_{t} \left|\theta_{i}(t)-\theta_{i}(t-1))||(R(t) - R(t-1))\right|.
\end{equation}

We then divide these two terms, plus a damping term $\zeta$, and take the absolute value:

\begin{equation}
\Omega^{k}_{i} = \left| \frac{\omega^{k}_{i}}{\overline{\omega}^{k}_{i} + \zeta} \right|.
\end{equation}

\subsection{Hyperparameter Search}

We tested our networks on different sets of hyperparameters to determine which values yielded the greatest mean classification accuracy. When using SI, we tested networks with with $c = \{0.001, 0.002, 0.005, 0.01, 0.02, 0.05,0.1, 0.2, 0.5, 1, 2\}$, and $\zeta = \{0.001, 0.01\}$ When using EWC, we tested networks with $c = \{0.1, 0.2, 0.5, 1, 2, 5, 10, 20, 100, 200, 500, 1000\}$. Furthermore, for the MNIST dataset, we tested networks with and without a 20\% drop rate in the input layer.  

Once the optimal $c$ and $\zeta$ (for SI) for each task were determined, we tested one additional value $c=\frac{c_{i}+c_{j}}{2}$, where $c_{i}$ and $c_{j}$ were the two values generating the greatest mean accuracies ($i$ and $j$ were always adjacent). The hyperparameters yielding the greatest mean classification accuracy across the 100 or 500 MNIST permutations, the 100 ImageNet tasks, or the 20 cognitive tasks, were used for Figures 2, 4, and 5.

Training the RNNs using reinforcement learning required additional hyperparameters, which we experimented with before settling on fixed values. We set the reward equal to $-1$ if the network broke fixation (i.e. did not choose the fixation action when required), equal to $-0.01$ when choosing the wrong output direction, and equal to $+1$ when choosing the correct output direction. The reward was $0$ for all other times, and the trial ended as soon as the network received a reward other than zero. The constant weighting the value loss function, $\beta$, was set to $0.01$. The discount factor, $\gamma$, was set to 0.9 although other values between 0 and 1 produced similar results. Lastly, the constant weighting the entropy loss function, $\alpha$ was set to 0.0001, and the learning rate was set to 0.0005.

\subsection{Analysis methods}

For our analysis into the interaction between synaptic stabilization and XdG, we trained two networks, one with stabilization (SI) and one with stabilization combined with XdG, on 100 sequential MNIST permutations. After training, we tested the network with SI alone after perturbing single connection weights located between the last hidden layer and the output layer. Specifically, we randomly selected 1000 connections weights one at a time, measured the mean accuracy on all 100 MNIST permutation after perturbing the weights by +10, and then by -10, and then compared the mean accuracy after perturbation to the mean accuracy without perturbation. Figure 3A shows the scatter plot of the difference in mean accuracy before and after perturbation (y-axis) with the synaptic importance (x-axis) calculated using equation 11.

For Figure 3B, we wanted to compare network flexibility (how much its connections weights and biases could be adjusted during training) and the network's accuracy on each new task.  Thus, we calculated the Euclidean distance between the values of the connection weights and biases before and after training on each task, and compared this value with the network's accuracy on the new task (Figure 3B). This was done for the entire sequence of 100 MNIST permutations.
 
For the comparison between synaptic importance and their change in value when learning a new task (Figures 3C-E, right panels), we binned the connection weights based on their importance using 80 evenly spaced bins on the logarithmic scale between 0.001 and 10 (same bins as used for histograms in left panels of the same figure). For display purposes, we set the minimum importance value to 0.001. Then, for all connection weights in a bin, we calculated their Euclidean distance before and after training on the 100th MNIST permutation. This was calculated for networks with SI only (magenta curves) and with SI combined with XdG (green curves).

\newpage
\bibliographystyle{unsrt}  
\bibliography{references}  %%% Use the external .bib file (using bibtex).

\begin{thebibliography}{10}

\bibitem{peters1991fine}
Alan Peters.
\newblock The fine structure of the nervous system.
\newblock {\em Neurons and their supporting cells}, 1991.

\bibitem{kasai2003structure}
Haruo Kasai, Masanori Matsuzaki, Jun Noguchi, Nobuaki Yasumatsu, and Hiroyuki
  Nakahara.
\newblock Structure--stability--function relationships of dendritic spines.
\newblock {\em Trends in neurosciences}, 26(7):360--368, 2003.

\bibitem{yuste2001morphological}
Rafael Yuste and Tobias Bonhoeffer.
\newblock Morphological changes in dendritic spines associated with long-term
  synaptic plasticity.
\newblock {\em Annual review of neuroscience}, 24(1):1071--1089, 2001.

\bibitem{yoshihara2009dendritic}
Yoshihiro Yoshihara, Mathias De~Roo, and Dominique Muller.
\newblock Dendritic spine formation and stabilization.
\newblock {\em Current opinion in neurobiology}, 19(2):146--153, 2009.

\bibitem{fischer1998rapid}
Maria Fischer, Stefanie Kaech, Darko Knutti, and Andrew Matus.
\newblock Rapid actin-based plasticity in dendritic spines.
\newblock {\em Neuron}, 20(5):847--854, 1998.

\bibitem{yang2009stably}
Guang Yang, Feng Pan, and Wen-Biao Gan.
\newblock Stably maintained dendritic spines are associated with lifelong
  memories.
\newblock {\em Nature}, 462(7275):920--924, 2009.

\bibitem{xu2009rapid}
Tonghui Xu, Xinzhu Yu, Andrew~J Perlik, Willie~F Tobin, Jonathan~A Zweig, Kelly
  Tennant, Theresa Jones, and Yi~Zuo.
\newblock Rapid formation and selective stabilization of synapses for enduring
  motor memories.
\newblock {\em Nature}, 462(7275):915--919, 2009.

\bibitem{zenke2017continual}
Friedemann Zenke, Ben Poole, and Surya Ganguli.
\newblock Continual learning through synaptic intelligence.
\newblock In {\em International Conference on Machine Learning}, pages
  3987--3995, 2017.

\bibitem{kirkpatrick2017overcoming}
James Kirkpatrick, Razvan Pascanu, Neil Rabinowitz, Joel Veness, Guillaume
  Desjardins, Andrei~A Rusu, Kieran Milan, John Quan, Tiago Ramalho, Agnieszka
  Grabska-Barwinska, et~al.
\newblock Overcoming catastrophic forgetting in neural networks.
\newblock {\em Proceedings of the National Academy of Sciences}, page
  201611835, 2017.

\bibitem{cichon2015branch}
Joseph Cichon and Wen-Biao Gan.
\newblock Branch-specific dendritic ca2+ spikes cause persistent synaptic
  plasticity.
\newblock {\em Nature}, 520(7546):180--185, 2015.

\bibitem{tononi2014sleep}
Giulio Tononi and Chiara Cirelli.
\newblock Sleep and the price of plasticity: from synaptic and cellular
  homeostasis to memory consolidation and integration.
\newblock {\em Neuron}, 81(1):12--34, 2014.

\bibitem{kukushkin2017memory}
Nikolay~Vadimovich Kukushkin and Thomas~James Carew.
\newblock Memory takes time.
\newblock {\em Neuron}, 95(2):259--279, 2017.

\bibitem{goodfellow2013empirical}
Ian~J Goodfellow, Mehdi Mirza, Da~Xiao, Aaron Courville, and Yoshua Bengio.
\newblock An empirical investigation of catastrophic forgetting in
  gradient-based neural networks.
\newblock {\em arXiv preprint arXiv:1312.6211}, 2013.

\bibitem{imagenet_cvpr09}
J.~Deng, W.~Dong, R.~Socher, L.-J. Li, K.~Li, and L.~Fei-Fei.
\newblock {ImageNet: A Large-Scale Hierarchical Image Database}.
\newblock In {\em CVPR09}, 2009.

\bibitem{yang2017clustering}
Guangyu~Robert Yang, H~Francis Song, William~T Newsome, and Xiao-Jing Wang.
\newblock Clustering and compositionality of task representations in a neural
  network trained to perform many cognitive tasks.
\newblock {\em bioRxiv}, page 183632, 2017.

\bibitem{engel2001dynamic}
Andreas~K Engel, Pascal Fries, and Wolf Singer.
\newblock Dynamic predictions: oscillations and synchrony in top--down
  processing.
\newblock {\em Nature Reviews Neuroscience}, 2(10):704--716, 2001.

\bibitem{johnston2007top}
Kevin Johnston, Helen~M Levin, Michael~J Koval, and Stefan Everling.
\newblock Top-down control-signal dynamics in anterior cingulate and prefrontal
  cortex neurons following task switching.
\newblock {\em Neuron}, 53(3):453--462, 2007.

\bibitem{miller2001integrative}
Earl~K Miller and Jonathan~D Cohen.
\newblock An integrative theory of prefrontal cortex function.
\newblock {\em Annual review of neuroscience}, 24(1):167--202, 2001.

\bibitem{kuchibhotla2017parallel}
Kishore~V Kuchibhotla, Jonathan~V Gill, Grace~W Lindsay, Eleni~S Papadoyannis,
  Rachel~E Field, Tom A~Hindmarsh Sten, Kenneth~D Miller, and Robert~C Froemke.
\newblock Parallel processing by cortical inhibition enables context-dependent
  behavior.
\newblock {\em Nature neuroscience}, 20(1):62--71, 2017.

\bibitem{otazu2009engaging}
Gonzalo~H Otazu, Lung-Hao Tai, Yang Yang, and Anthony~M Zador.
\newblock Engaging in an auditory task suppresses responses in auditory cortex.
\newblock {\em Nature neuroscience}, 12(5):646--654, 2009.

\bibitem{hochreiter1997long}
Sepp Hochreiter and J{\"u}rgen Schmidhuber.
\newblock Long short-term memory.
\newblock {\em Neural computation}, 9(8):1735--1780, 1997.

\bibitem{santoro2016one}
Adam Santoro, Sergey Bartunov, Matthew Botvinick, Daan Wierstra, and Timothy
  Lillicrap.
\newblock One-shot learning with memory-augmented neural networks.
\newblock {\em arXiv preprint arXiv:1605.06065}, 2016.

\bibitem{lake2013one}
Brenden~M Lake, Ruslan~R Salakhutdinov, and Josh Tenenbaum.
\newblock One-shot learning by inverting a compositional causal process.
\newblock In {\em Advances in neural information processing systems}, pages
  2526--2534, 2013.

\bibitem{fernando2017pathnet}
Chrisantha Fernando, Dylan Banarse, Charles Blundell, Yori Zwols, David Ha,
  Andrei~A Rusu, Alexander Pritzel, and Daan Wierstra.
\newblock Pathnet: Evolution channels gradient descent in super neural
  networks.
\newblock {\em arXiv preprint arXiv:1701.08734}, 2017.

\bibitem{bassett2011dynamic}
Danielle~S Bassett, Nicholas~F Wymbs, Mason~A Porter, Peter~J Mucha, Jean~M
  Carlson, and Scott~T Grafton.
\newblock Dynamic reconfiguration of human brain networks during learning.
\newblock {\em Proceedings of the National Academy of Sciences}, 2011.

\bibitem{velez2017diffusion}
Roby Velez and Jeff Clune.
\newblock Diffusion-based neuromodulation can eliminate catastrophic forgetting
  in simple neural networks.
\newblock {\em arXiv preprint arXiv:1705.07241}, 2017.

\bibitem{rusu2016progressive}
Andrei~A Rusu, Neil~C Rabinowitz, Guillaume Desjardins, Hubert Soyer, James
  Kirkpatrick, Koray Kavukcuoglu, Razvan Pascanu, and Raia Hadsell.
\newblock Progressive neural networks.
\newblock {\em arXiv preprint arXiv:1606.04671}, 2016.

\bibitem{li2017learning}
Zhizhong Li and Derek Hoiem.
\newblock Learning without forgetting.
\newblock {\em IEEE Transactions on Pattern Analysis and Machine Intelligence},
  2017.

\bibitem{aljundi2017memory}
Rahaf Aljundi, Francesca Babiloni, Mohamed Elhoseiny, Marcus Rohrbach, and
  Tinne Tuytelaars.
\newblock Memory aware synapses: Learning what (not) to forget.
\newblock {\em arXiv preprint arXiv:1711.09601}, 2017.

\bibitem{nguyen2017variational}
Cuong~V Nguyen, Yingzhen Li, Thang~D Bui, and Richard~E Turner.
\newblock Variational continual learning.
\newblock {\em arXiv preprint arXiv:1710.10628}, 2017.

\bibitem{he2017overcoming}
Xu~He and Herbert Jaeger.
\newblock Overcoming catastrophic interference by conceptors.
\newblock {\em arXiv preprint arXiv:1707.04853}, 2017.

\bibitem{mallya2017packnet}
Arun Mallya and Svetlana Lazebnik.
\newblock Packnet: Adding multiple tasks to a single network by iterative
  pruning.
\newblock {\em arXiv preprint arXiv:1711.05769}, 2017.

\bibitem{serra2018overcoming}
Joan Serr{\`a}, D{\'\i}dac Sur{\'\i}s, Marius Miron, and Alexandros
  Karatzoglou.
\newblock Overcoming catastrophic forgetting with hard attention to the task.
\newblock {\em arXiv preprint arXiv:1801.01423}, 2018.

\bibitem{abadi2016tensorflow}
Mart{\'\i}n Abadi, Ashish Agarwal, Paul Barham, Eugene Brevdo, Zhifeng Chen,
  Craig Citro, Greg~S Corrado, Andy Davis, Jeffrey Dean, Matthieu Devin, et~al.
\newblock Tensorflow: Large-scale machine learning on heterogeneous distributed
  systems.
\newblock {\em arXiv preprint arXiv:1603.04467}, 2016.

\bibitem{srivastava2014dropout}
Nitish Srivastava, Geoffrey Hinton, Alex Krizhevsky, Ilya Sutskever, and Ruslan
  Salakhutdinov.
\newblock Dropout: A simple way to prevent neural networks from overfitting.
\newblock {\em The Journal of Machine Learning Research}, 15(1):1929--1958,
  2014.

\bibitem{kingma2014adam}
Diederik Kingma and Jimmy Ba.
\newblock Adam: A method for stochastic optimization.
\newblock {\em arXiv preprint arXiv:1412.6980}, 2014.

\bibitem{barto1983neuronlike}
Andrew~G Barto, Richard~S Sutton, and Charles~W Anderson.
\newblock Neuronlike adaptive elements that can solve difficult learning
  control problems.
\newblock {\em IEEE transactions on systems, man, and cybernetics},
  (5):834--846, 1983.

\bibitem{schulman2015high}
John Schulman, Philipp Moritz, Sergey Levine, Michael Jordan, and Pieter
  Abbeel.
\newblock High-dimensional continuous control using generalized advantage
  estimation.
\newblock {\em arXiv preprint arXiv:1506.02438}, 2015.

\end{thebibliography}
\newpage

%%%%%%%%%%%%%%%%%%%%%%%%%%%%%%%%%%%%%%%%%%%%%%%%%%%%%%%%%%%%%%%%%%%%%%%%

\setcounter{figure}{0}
\makeatletter
\renewcommand\thefigure{S\@arabic\c@figure}
\makeatother

\makeatletter
\renewcommand\thetable{S\@arabic\c@table}
\makeatother

\section*{Supplemental Information}

\subsection*{Network Tasks Overview}

We trained RNNs on a set of 20 cognition-based tasks, commonly used in neuroscience research. These 20 tasks were implemented in a manner similar to \cite{yang2017clustering}. In all 20 tasks, stimuli were represented as coherent motion patterns moving in one of eight, equally spaced, possible directions. However, the results are not meant to be specific to motion, or even visual, inputs. The network output represents a decision to either withhold a response (in many neuroscience experiments, this is equivalent to maintaining fixation), or to respond in one of eight different directions (equivalent to generating a saccade).

Stimuli could be presented in one of two locations; neural responses to stimuli in one location were not influenced by stimuli in the other location. A "fixation" cue was also presented, which usually (but not always) indicated to the network to without a response. The fixation cue was always presented at the start of each trial, and lasted for a variable amount of time depending on the task contingencies. All tasks lasted for 2000 ms, with time steps of 20 ms. For some networks (green dots, Figure 5B) we also included a rule cue which indicated to the network which task was currently active.

The twenty tasks can be broken down in five different groups: Go, Anti-Go, Decision-Making (DM), Delayed Decision-Making (Dly DM), and Matching.

\subsection*{Go Group Tasks}

The Go group includes the Go, Reaction Time Go, and the Delay Go tasks (abbreviated as Go, RT Go, and Dly Go, respectively). In all three tasks, a single stimulus is presented, and the network is trained to respond in the direction of that stimulus.  

In the Go task, the stimulus appears at a random time between 400 and 1400 ms (uniform distribution) and lasted until the end of the trial. The fixation turned off at 1400 ms, at which time the network could respond.

The RT-Go task was similar to the Go task except that the fixation cue never turned off. The network could respond as soon as the motion stimulus appeared.

In the Dly task, the stimulus was presented from 400 to 700 ms, and the fixation cue was turned off at either 900, 1100 or 1500 ms, at which time the network could respond.

\subsection*{Anti-Go Group Tasks}

The Anti-Go group includes the the Anti-Go, Anti-Reaction Time Go, and the Anti-Delay Go tasks (abbreviated as Anti-Go, Anti-RT Go, and Anti-Dly Go, respectively). These tasks were similar to the tasks from the Go group, except that the network was trained to respond to the direction 180 degrees opposite to the presented motion direction. 

\subsection*{Decision-Making Group Tasks}

The Decision-Making group includes the Decision-Making 1 and 2, Context-Dependent Decision-Making 1 and 2, and Multi-Stimulus Decision Making tasks (abbreviated as DM1, DM2, Ctx DM1, Ctx DM2, and MultStim DM tasks, respectively). In these tasks, two motion directions were randomly chosen such that they were always separated by between $\pi/2$ and $3\pi/2$. These two motion directions were either presented simultaneously at a single location, or both directions were presented in both locations. The strength of the motion direction stimuli were set to $\gamma_{1, i} = \bar\gamma + c_{i}$, and $\gamma_{2, i} = \bar\gamma - c_{i}$, where $\bar\gamma$ was sampled from the uniform distribution $(0.8, 1.2)$ on a trial-by-trial basis, and difference in strength between the two stimuli, $c_{i}$, was randomly selected from the set $\{0.1, 0.2, 0.4\}$. Stimuli were presented from 400 ms after trial start and lasted either 200, 400 or 800 ms. The fixation cue was turned off as soon as the stimuli were extinguished. 

In the DM 1 and DM2 tasks, two stimuli were presented in either the first or second location, respectively, and the network was trained to respond towards the direction associated with the stronger motion stimulus. In the Ctx DM1 and Ctx DM2 tasks, pairs of motion directions were presented in both locations, but only stimuli in location 1 were task-relevant for Ctx DM1, and only stimuli in location 2 were task-relevant for Ctx DM2. In the MultStim DM task, the same two motion directions were presented in both locations, although their strengths varied. The network was had to integrate information from both locations to generate the correct response. Specifically, it was trained to respond to the direction associated with the motion direction with the greater mean strength across both locations.

\subsection*{Delay Decision-Making Group Tasks}

The Delay Decision-Making group includes Delay Decision-Making 1 and 2, Context-Dependent Delay Decision-Making 1 and 2, and Multi-Stimulus Delay Decision Making tasks (abbreviated as Dly DM1, Dly Dm2, Ctx Dly DM1, Ctx Dly DM2 and MultStim Dly DM tasks, respectively). These tasks were similar to the tasks in the Decision-Making group except the timing of the stimulus presentation and their relative strengths. In the Delay group, instead of presenting both motion directions simultaneously, both motion directions were presented for 300 ms, with the first presented at 400 ms after trial start, and the second presented at either 200, 400 or 800 ms after the first stimulus was extinguished.  Also, the difference in strength between the two stimuli, $c_{i}$, was randomly selected from $\{0.1, 0.2, 0.4\}$.

\subsection*{Matching Group Tasks}

The Matching group includes the Delayed Match-to-Sample, Anti-Delayed Match-to-Sample, Delayed Match-to-Category, and the Anti-Delayed Match-to-Category (abbreviated as DMS, ADMS, DMC, ADMC tasks, respectively). In all four tasks, two 300 ms motion direction stimuli are presented in either location 1 or 2. The first stimulus was presented 400 ms after trial start, and the second was presented either 200, 400 or 800 ms after the first stimulus was extinguished. The fixation cue was turned off as soon as the second stimulus was extinguished.
 
In the DMS and ADMS tasks, the stimuli were considered a match if the motion direction of the two stimuli were identical. The network was to respond only if the two stimuli matched (which occurred for 50\% of trials). In the DMS task, the network was trained to respond towards the direction of the matching motion direction, and in the ADMS task, the network was trained to respond towards the direction 180 degrees away from the matching direction.

In the DMC and ADMC tasks, the eight motion directions were divided into two categories, each containing four adjacent directions). The two stimuli were considered a match if the two motion directions belonged the same category (which occurred for 50\% of trials). The network was to respond only if the two stimuli matched. In the DMC task, the network was trained to respond towards the second of the two motion directions, and in the ADMS task, the network was trained to respond towards the direction 180 degrees away from the second motion direction.

\vspace{2cm}
\begin{figure}[h]
\centering
\includegraphics[width=0.5\textwidth]{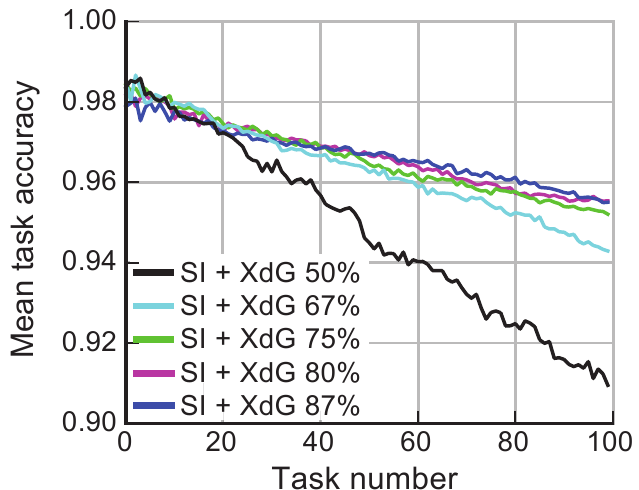}
\caption{\small Mean classification accuracy for networks on the permuted MNIST task.  All curves use SI combined with context-dependent gating.  The percentages of gated hidden units per task are 50\% (black curve), 67\%  (light blue curve), 75\% (green curve), 80\%  (magenta curve), and 86.7\% (dark blue curve).  Mean classification accuracy for our network architecture trained across 100 permutations peaks when the percentage of gated hidden units per task is between 80\% and 86.7\%.}
\label{fig:figS1}
\end{figure}

\begin{figure}
\centering
\includegraphics[width=0.5\linewidth]{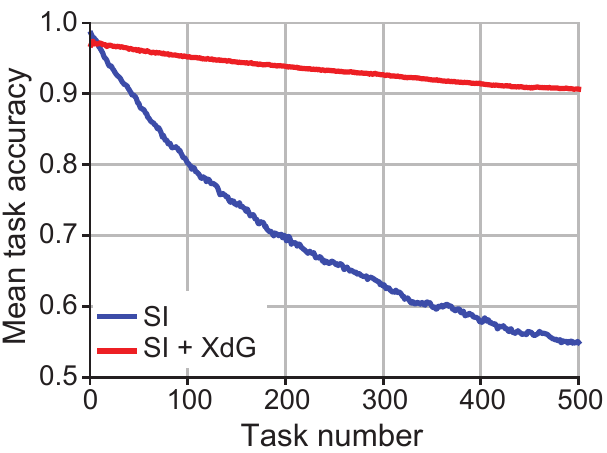}
\caption{\small Mean classification accuracy across 500 permutations of the MNIST task. The blue curve shows the mean classification accuracy for networks with SI, and the red curve shows the mean accuracy for networks with SI combined with context-dependent gating in which 92.2\% (11/12) of units were gated for each task.}
\label{fig:figS2}
\end{figure}

\begin{figure}
\centering
\includegraphics[width=0.5\linewidth]{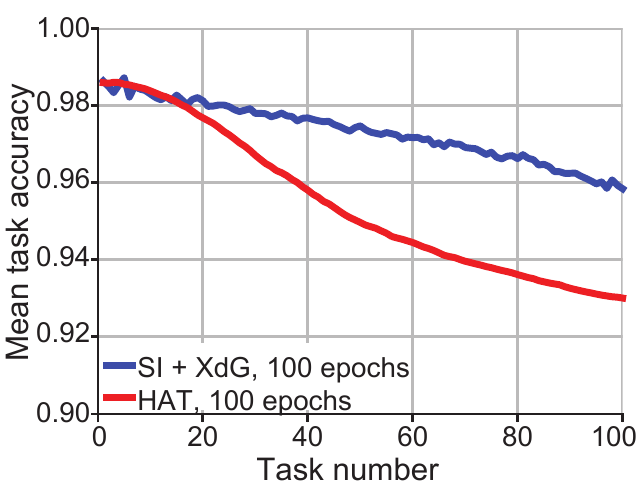}
\caption{\small Comparing the Hard Attention to Task (HAT) method \cite{serra2018overcoming} to XdG + parameter stabilization (SI). Both networks were trained across 100 MNIST permutations, and the mean accuracy of all previously trained tasks is shown. Both methods were trained for 100 epochs on each task (compared to 20 epochs that we used in Figure 2). The code used to run the HAT method was taken code that accompanied the paper \href{https://github.com/joansj/hat}{https://github.com/joansj/hat}. We set $\lambda = 0.75$ which appeared to provide maximum mean accuracy. All other hyperparameters aside from the number of training epochs and $\lambda$ were kept at the default values found in the code provided. Mean accuracy across all 100 permutation was 95.8\% for XdG combined with stabilization, and 93.0\% for HAT.}
\label{fig:figS3}
\end{figure}

\newpage

\begin{table}[p]
	\centering
    \caption{ImageNet labels assigned to each ImageNet classification task.}
    \label{table:Table1}
	\begin{tabular}{|c|p{14cm}|}
		
		\hline
        \textbf{Task} & \textbf{Labels in Task} \\ \hline
		\scriptsize 0 & \scriptsize kit fox, English setter, Siberian husky, Australian terrier, English springer, grey whale, lesser panda, Egyptian cat, ibex, Persian cat \\ \hline
		\scriptsize 1 & \scriptsize cougar, gazelle, porcupine, sea lion, malamute, badger, Great Dane, Walker hound, Welsh springer spaniel, whippet \\ \hline
		\scriptsize 2 & \scriptsize Scottish deerhound, killer whale, mink, African elephant, Weimaraner, soft-coated wheaten terrier, Dandie Dinmont, red wolf, Old English sheepdog, jaguar \\ \hline
		\scriptsize 3 & \scriptsize otterhound, bloodhound, Airedale, hyena, meerkat, giant schnauzer, titi, three-toed sloth, sorrel, black-footed ferret \\ \hline
		\scriptsize 4 & \scriptsize dalmatian, black-and-tan coonhound, papillon, skunk, Staffordshire bullterrier, Mexican hairless, Bouvier des Flandres, weasel, miniature poodle, Cardigan \\ \hline
		\scriptsize 5 & \scriptsize malinois, bighorn, fox squirrel, colobus, tiger cat, Lhasa, impala, coyote, Yorkshire terrier, Newfoundland \\ \hline
		\scriptsize 6 & \scriptsize brown bear, red fox, Norwegian elkhound, Rottweiler, hartebeest, Saluki, grey fox, schipperke, Pekinese, Brabancon griffon \\ \hline
		\scriptsize 7 & \scriptsize West Highland white terrier, Sealyham terrier, guenon, mongoose, indri, tiger, Irish wolfhound, wild boar, EntleBucher, zebra \\ \hline
		\scriptsize 8 & \scriptsize ram, French bulldog, orangutan, basenji, leopard, Bernese mountain dog, Maltese dog, Norfolk terrier, toy terrier, vizsla \\ \hline
		\scriptsize 9 & \scriptsize cairn, squirrel monkey, groenendael, clumber, Siamese cat, chimpanzee, komondor, Afghan hound, Japanese spaniel, proboscis monkey \\ \hline
		\scriptsize 10 & \scriptsize guinea pig, white wolf, ice bear, gorilla, borzoi, toy poodle, Kerry blue terrier, ox, Scotch terrier, Tibetan mastiff \\ \hline
		\scriptsize 11 & \scriptsize spider monkey, Doberman, Boston bull, Greater Swiss Mountain dog, Appenzeller, Shih-Tzu, Irish water spaniel, Pomeranian, Bedlington terrier, warthog \\ \hline
		\scriptsize 12 & \scriptsize Arabian camel, siamang, miniature schnauzer, collie, golden retriever, Irish terrier, affenpinscher, Border collie, hare, boxer \\ \hline
		\scriptsize 13 & \scriptsize silky terrier, beagle, Leonberg, German short-haired pointer, patas, dhole, baboon, macaque, Chesapeake Bay retriever, bull mastiff \\ \hline
		\scriptsize 14 & \scriptsize kuvasz, capuchin, pug, curly-coated retriever, Norwich terrier, flat-coated retriever, hog, keeshond, Eskimo dog, Brittany spaniel \\ \hline
		\scriptsize 15 & \scriptsize standard poodle, Lakeland terrier, snow leopard, Gordon setter, dingo, standard schnauzer, hamster, Tibetan terrier, Arctic fox, wire-haired fox terrier \\ \hline
		\scriptsize 16 & \scriptsize basset, water buffalo, American black bear, Angora, bison, howler monkey, hippopotamus, chow, giant panda, American Staffordshire terrier \\ \hline
		\scriptsize 17 & \scriptsize Shetland sheepdog, Great Pyrenees, Chihuahua, tabby, marmoset, Labrador retriever, Saint Bernard, armadillo, Samoyed, bluetick \\ \hline
		\scriptsize 18 & \scriptsize redbone, polecat, marmot, kelpie, gibbon, llama, miniature pinscher, wood rabbit, Italian greyhound, lion \\ \hline
		\scriptsize 19 & \scriptsize cocker spaniel, Irish setter, dugong, Indian elephant, beaver, Sussex spaniel, Pembroke, Blenheim spaniel, Madagascar cat, Rhodesian ridgeback \\ \hline
		\scriptsize 20 & \scriptsize lynx, African hunting dog, langur, Ibizan hound, timber wolf, cheetah, English foxhound, briard, sloth bear, Border terrier \\ \hline
		\scriptsize 21 & \scriptsize German shepherd, otter, koala, tusker, echidna, wallaby, platypus, wombat, revolver, umbrella \\ \hline
		\scriptsize 22 & \scriptsize schooner, soccer ball, accordion, ant, starfish, chambered nautilus, grand piano, laptop, strawberry, airliner \\ \hline
		\scriptsize 23 & \scriptsize warplane, airship, balloon, space shuttle, fireboat, gondola, speedboat, lifeboat, canoe, yawl \\ \hline
		\scriptsize 24 & \scriptsize catamaran, trimaran, container ship, liner, pirate, aircraft carrier, submarine, wreck, half track, tank \\ \hline
		\scriptsize 25 & \scriptsize missile, bobsled, dogsled, bicycle-built-for-two, mountain bike, freight car, passenger car, barrow, shopping cart, motor scooter \\ \hline
		\scriptsize 26 & \scriptsize forklift, electric locomotive, steam locomotive, amphibian, ambulance, beach wagon, cab, convertible, jeep, limousine \\ \hline
		\scriptsize 27 & \scriptsize minivan, Model T, racer, sports car, go-kart, golfcart, moped, snowplow, fire engine, garbage truck \\ \hline
		\scriptsize 28 & \scriptsize pickup, tow truck, trailer truck, moving van, police van, recreational vehicle, streetcar, snowmobile, tractor, mobile home \\ \hline
		\scriptsize 29 & \scriptsize tricycle, unicycle, horse cart, jinrikisha, oxcart, bassinet, cradle, crib, four-poster, bookcase \\ \hline
		\scriptsize 30 & \scriptsize china cabinet, medicine chest, chiffonier, table lamp, file, park bench, barber chair, throne, folding chair, rocking chair \\ \hline
		\scriptsize 31 & \scriptsize studio couch, toilet seat, desk, pool table, dining table, entertainment center, wardrobe, Granny Smith, orange, lemon \\ \hline
		\scriptsize 32 & \scriptsize fig, pineapple, banana, jackfruit, custard apple, pomegranate, acorn, hip, ear, rapeseed \\ \hline
		\scriptsize 33 & \scriptsize corn, buckeye, organ, upright, chime, drum, gong, maraca, marimba, steel drum \\ \hline
		\scriptsize 34 & \scriptsize banjo, cello, violin, harp, acoustic guitar, electric guitar, cornet, French horn, trombone, harmonica \\ \hline
		\scriptsize 35 & \scriptsize ocarina, panpipe, bassoon, oboe, sax, flute, daisy, yellow lady's slipper, cliff, valley \\ \hline
		\scriptsize 36 & \scriptsize alp, volcano, promontory, sandbar, coral reef, lakeside, seashore, geyser, hatchet, cleaver \\ \hline
		\scriptsize 37 & \scriptsize letter opener, plane, power drill, lawn mower, hammer, corkscrew, can opener, plunger, screwdriver, shovel \\ \hline
		\scriptsize 38 & \scriptsize plow, chain saw, cock, hen, ostrich, brambling, goldfinch, house finch, junco, indigo bunting \\ \hline
		\scriptsize 39 & \scriptsize robin, bulbul, jay, magpie, chickadee, water ouzel, kite, bald eagle, vulture, great grey owl \\ \hline
		\scriptsize 40 & \scriptsize black grouse, ptarmigan, ruffed grouse, prairie chicken, peacock, quail, partridge, African grey, macaw, sulphur-crested cockatoo \\ \hline
		\scriptsize 41 & \scriptsize lorikeet, coucal, bee eater, hornbill, hummingbird, jacamar, toucan, drake, red-breasted merganser, goose \\ \hline
		\scriptsize 42 & \scriptsize black swan, white stork, black stork, spoonbill, flamingo, American egret, little blue heron, bittern, crane, limpkin \\ \hline
		\scriptsize 43 & \scriptsize American coot, bustard, ruddy turnstone, red-backed sandpiper, redshank, dowitcher, oystercatcher, European gallinule, pelican, king penguin \\ \hline
		\scriptsize 44 & \scriptsize albatross, great white shark, tiger shark, hammerhead, electric ray, stingray, barracouta, coho, tench, goldfish \\ \hline
		\scriptsize 45 & \scriptsize eel, rock beauty, anemone fish, lionfish, puffer, sturgeon, gar, loggerhead, leatherback turtle, mud turtle \\ \hline
		\scriptsize 46 & \scriptsize terrapin, box turtle, banded gecko, common iguana, American chameleon, whiptail, agama, frilled lizard, alligator lizard, Gila monster \\ \hline
		\scriptsize 47 & \scriptsize green lizard, African chameleon, Komodo dragon, triceratops, African crocodile, American alligator, thunder snake, ringneck snake, hognose snake, green snake \\ \hline
		\scriptsize 48 & \scriptsize king snake, garter snake, water snake, vine snake, night snake, boa constrictor, rock python, Indian cobra, green mamba, sea snake \\ \hline
		\scriptsize 49 & \scriptsize horned viper, diamondback, sidewinder, European fire salamander, common newt, eft, spotted salamander, axolotl, bullfrog, tree frog \\ \hline
		
	\end{tabular}	
\end{table}	

\begin{table}[p]
	\centering
		\begin{tabular}{|c|p{14cm}|}
		
		\hline
		\scriptsize 50 & \scriptsize tailed frog, whistle, wing, paintbrush, hand blower, oxygen mask, snorkel, loudspeaker, microphone, screen \\ \hline
		\scriptsize 51 & \scriptsize mouse, electric fan, oil filter, strainer, space heater, stove, guillotine, barometer, rule, odometer \\ \hline
		\scriptsize 52 & \scriptsize scale, analog clock, digital clock, wall clock, hourglass, sundial, parking meter, stopwatch, digital watch, stethoscope \\ \hline
		\scriptsize 53 & \scriptsize syringe, magnetic compass, binoculars, projector, sunglasses, loupe, radio telescope, bow, cannon, assault rifle \\ \hline
		\scriptsize 54 & \scriptsize rifle, projectile, computer keyboard, typewriter keyboard, crane, lighter, abacus, cash machine, slide rule, desktop computer \\ \hline
		\scriptsize 55 & \scriptsize hand-held computer, notebook, web site, harvester, thresher, printer, slot, vending machine, sewing machine, joystick \\ \hline
		\scriptsize 56 & \scriptsize switch, hook, car wheel, paddlewheel, pinwheel, potter's wheel, gas pump, carousel, swing, reel \\ \hline
		\scriptsize 57 & \scriptsize radiator, puck, hard disc, sunglass, pick, car mirror, solar dish, remote control, disk brake, buckle \\ \hline
		\scriptsize 58 & \scriptsize hair slide, knot, combination lock, padlock, nail, safety pin, screw, muzzle, seat belt, ski \\ \hline
		\scriptsize 59 & \scriptsize candle, jack-o'-lantern, spotlight, torch, neck brace, pier, tripod, maypole, mousetrap, spider web \\ \hline
		\scriptsize 60 & \scriptsize trilobite, harvestman, scorpion, black and gold garden spider, barn spider, garden spider, black widow, tarantula, wolf spider, tick \\ \hline
		\scriptsize 61 & \scriptsize centipede, isopod, Dungeness crab, rock crab, fiddler crab, king crab, American lobster, spiny lobster, crayfish, hermit crab \\ \hline
		\scriptsize 62 & \scriptsize tiger beetle, ladybug, ground beetle, long-horned beetle, leaf beetle, dung beetle, rhinoceros beetle, weevil, fly, bee \\ \hline
		\scriptsize 63 & \scriptsize grasshopper, cricket, walking stick, cockroach, mantis, cicada, leafhopper, lacewing, dragonfly, damselfly \\ \hline
		\scriptsize 64 & \scriptsize admiral, ringlet, monarch, cabbage butterfly, sulphur butterfly, lycaenid, jellyfish, sea anemone, brain coral, flatworm \\ \hline
		\scriptsize 65 & \scriptsize nematode, conch, snail, slug, sea slug, chiton, sea urchin, sea cucumber, iron, espresso maker \\ \hline
		\scriptsize 66 & \scriptsize microwave, Dutch oven, rotisserie, toaster, waffle iron, vacuum, dishwasher, refrigerator, washer, Crock Pot \\ \hline
		\scriptsize 67 & \scriptsize frying pan, wok, caldron, coffeepot, teapot, spatula, altar, triumphal arch, patio, steel arch bridge \\ \hline
		\scriptsize 68 & \scriptsize suspension bridge, viaduct, barn, greenhouse, palace, monastery, library, apiary, boathouse, church \\ \hline
		\scriptsize 69 & \scriptsize mosque, stupa, planetarium, restaurant, cinema, home theater, lumbermill, coil, obelisk, totem pole \\ \hline
		\scriptsize 70 & \scriptsize castle, prison, grocery store, bakery, barbershop, bookshop, butcher shop, confectionery, shoe shop, tobacco shop \\ \hline
		\scriptsize 71 & \scriptsize toyshop, fountain, cliff dwelling, yurt, dock, brass, megalith, bannister, breakwater, dam \\ \hline
		\scriptsize 72 & \scriptsize chainlink fence, picket fence, worm fence, stone wall, grille, sliding door, turnstile, mountain tent, scoreboard, honeycomb \\ \hline
		\scriptsize 73 & \scriptsize plate rack, pedestal, beacon, mashed potato, bell pepper, head cabbage, broccoli, cauliflower, zucchini, spaghetti squash \\ \hline
		\scriptsize 74 & \scriptsize acorn squash, butternut squash, cucumber, artichoke, cardoon, mushroom, shower curtain, jean, carton, handkerchief \\ \hline
		\scriptsize 75 & \scriptsize sandal, ashcan, safe, plate, necklace, croquet ball, fur coat, thimble, pajama, running shoe \\ \hline
		\scriptsize 76 & \scriptsize cocktail shaker, chest, manhole cover, modem, tub, tray, balance beam, bagel, prayer rug, kimono \\ \hline
		\scriptsize 77 & \scriptsize hot pot, whiskey jug, knee pad, book jacket, spindle, ski mask, beer bottle, crash helmet, bottlecap, tile roof \\ \hline
		\scriptsize 78 & \scriptsize mask, maillot, Petri dish, football helmet, bathing cap, teddy, holster, pop bottle, photocopier, vestment \\ \hline
		\scriptsize 79 & \scriptsize crossword puzzle, golf ball, trifle, suit, water tower, feather boa, cloak, red wine, drumstick, shield \\ \hline
		\scriptsize 80 & \scriptsize Christmas stocking, hoopskirt, menu, stage, bonnet, meat loaf, baseball, face powder, scabbard, sunscreen \\ \hline
		\scriptsize 81 & \scriptsize beer glass, hen-of-the-woods, guacamole, lampshade, wool, hay, bow tie, mailbag, water jug, bucket \\ \hline
		\scriptsize 82 & \scriptsize dishrag, soup bowl, eggnog, mortar, trench coat, paddle, chain, swab, mixing bowl, potpie \\ \hline
		\scriptsize 83 & \scriptsize wine bottle, shoji, bulletproof vest, drilling platform, binder, cardigan, sweatshirt, pot, birdhouse, hamper \\ \hline
		\scriptsize 84 & \scriptsize ping-pong ball, pencil box, pay-phone, consomme, apron, punching bag, backpack, groom, bearskin, pencil sharpener \\ \hline
		\scriptsize 85 & \scriptsize broom, mosquito net, abaya, mortarboard, poncho, crutch, Polaroid camera, space bar, cup, racket \\ \hline
		\scriptsize 86 & \scriptsize traffic light, quill, radio, dough, cuirass, military uniform, lipstick, shower cap, monitor, oscilloscope \\ \hline
		\scriptsize 87 & \scriptsize mitten, brassiere, French loaf, vase, milk can, rugby ball, paper towel, earthstar, envelope, miniskirt \\ \hline
		\scriptsize 88 & \scriptsize cowboy hat, trolleybus, perfume, bathtub, hotdog, coral fungus, bullet train, pillow, toilet tissue, cassette \\ \hline
		\scriptsize 89 & \scriptsize carpenter's kit, ladle, stinkhorn, lotion, hair spray, academic gown, dome, crate, wig, burrito \\ \hline
		\scriptsize 90 & \scriptsize pill bottle, chain mail, theater curtain, window shade, barrel, washbasin, ballpoint, basketball, bath towel, cowboy boot \\ \hline
		\scriptsize 91 & \scriptsize gown, window screen, agaric, cellular telephone, nipple, barbell, mailbox, lab coat, fire screen, minibus \\ \hline
		\scriptsize 92 & \scriptsize packet, maze, pole, horizontal bar, sombrero, pickelhaube, rain barrel, wallet, cassette player, comic book \\ \hline
		\scriptsize 93 & \scriptsize piggy bank, street sign, bell cote, fountain pen, Windsor tie, volleyball, overskirt, sarong, purse, bolo tie \\ \hline
		\scriptsize 94 & \scriptsize bib, parachute, sleeping bag, television, swimming trunks, measuring cup, espresso, pizza, breastplate, shopping basket \\ \hline
		\scriptsize 95 & \scriptsize wooden spoon, saltshaker, chocolate sauce, ballplayer, goblet, gyromitra, stretcher, water bottle, dial telephone, soap dispenser \\ \hline
		\scriptsize 96 & \scriptsize jersey, school bus, jigsaw puzzle, plastic bag, reflex camera, diaper, Band Aid, ice lolly, velvet, tennis ball \\ \hline
		\scriptsize 97 & \scriptsize gasmask, doormat, Loafer, ice cream, pretzel, quilt, maillot, tape player, clog, iPod \\ \hline
		\scriptsize 98 & \scriptsize bolete, scuba diver, pitcher, matchstick, bikini, sock, CD player, lens cap, thatch, vault \\ \hline
		\scriptsize 99 & \scriptsize beaker, bubble, cheeseburger, parallel bars, flagpole, coffee mug, rubber eraser, stole, carbonara, dumbbell \\ \hline
		
	\end{tabular}	
\end{table}

\end{document}